\documentclass[letterpaper]{article} 
\usepackage{aaai24}  
\usepackage{times}  
\usepackage{helvet}  
\usepackage{courier}  
\usepackage[hyphens]{url}  
\usepackage{graphicx} 
\usepackage{natbib}  
\usepackage{caption} 
\usepackage{colortbl}
\usepackage{subfigure}
\usepackage{savesym}
\usepackage[ruled,vlined,linesnumbered]{algorithm2e}
\savesymbol{listofalgorithms}

\usepackage{algorithm}
\restoresymbol{algo}{listofalgorithms}
\usepackage{algorithmic}
\usepackage{microtype}      
\usepackage{xcolor}         
\usepackage{multirow}
\usepackage[switch]{lineno}
\usepackage{booktabs}       
\usepackage{amsfonts}       
\usepackage{nicefrac}       
\usepackage{newfloat}
\usepackage{listings}
\DeclareCaptionStyle{ruled}{labelfont=normalfont,labelsep=colon,strut=off} 
\floatstyle{ruled}
\newfloat{listing}{tb}{lst}{}
\floatname{listing}{Listing}
\newcommand{\modelname}{TiMix }
\newcommand{\PretrainTaskName}{Patch Text Alignment }
\usepackage{amsmath}
\title{TiMix: Text-Aware Image Mixing for Effective Vision-Language Pre-training}

\author {
    Chaoya Jiang\textsuperscript{\rm 1},
    Wei Ye\textsuperscript{\rm 1}\thanks{~~Corresponding Author.},
    Haiyang Xu\textsuperscript{\rm 2}\footnotemark[1],
    Qinghao Ye\textsuperscript{\rm 2},\\
    Ming Yan\textsuperscript{\rm 2},
    Ji Zhang\textsuperscript{\rm 2},
    Shikun Zhang\textsuperscript{\rm 1}
}
\affiliations {
    \textsuperscript{\rm 1}National Engineering Research Center for Software Engineering, Peking University, Beijing, China\\
    \textsuperscript{\rm 2}Alibaba Group, Hangzhou, China\\
    wye@pku.edu.cn, shuofeng.xhy@alibaba-inc.com
}

\usepackage{bibentry}

\begin{document}

\maketitle

\begin{abstract}
Self-supervised Multi-modal Contrastive Learning (SMCL) remarkably advances modern Vision-Language Pre-training (VLP) models by aligning visual and linguistic modalities. Due to noises in web-harvested text-image pairs, however, scaling up training data volume in SMCL presents considerable obstacles in terms of computational cost and data inefficiency. To improve data efficiency in VLP, we propose Text-aware Image Mixing (\modelname), which integrates mix-based data augmentation techniques into SMCL, yielding significant performance improvements without significantly increasing computational overhead. We provide a theoretical analysis of \modelname from a mutual information (MI) perspective, showing that mixed data samples for cross-modal contrastive learning implicitly serve as a regularizer for the contrastive loss. The experimental results demonstrate that \modelname exhibits a comparable performance on downstream tasks, even with a reduced amount of training data and shorter training time, when benchmarked against existing methods. This work empirically and theoretically demonstrates the potential of data mixing for data-efficient and computationally viable VLP, benefiting broader VLP model adoption in practical scenarios. Our code is available on \color{blue}{\url{https://github.com/chaoyajiang/TiMiX/tree/main}}.
\end{abstract}

\section*{Introduction}

Vision-Language Pre-training (VLP) exploits large-scale image-text pairs without annotations via self-supervised learning~\cite{Syed2021ASO,Liu2020SelfSupervisedLG}, achieving tremendous success on a wide range of cross-modal downstream tasks~\cite{Tan2019LXMERTLC,Chen2020UNITERUI,Huang2020PixelBERTAI,Li2020OscarOA,Jiang2023BUS,Jiang2023COPA,Jiang2023SACL,Yu2021ERNIEViLKE,Li2021AlignBF,Wang2021SimVLMSV,Alayrac2022FlamingoAV,2021VinVL,Kim2021ViLTVT,li2022mplug,Xu2021E2EVLPEV}. More recently, Self-supervised Multi-modal Contrastive Learning (SMCL)  has emerged as a significant advancement in the VLP community~\cite{li2022blip,clip, Li2021AlignBF, li2022mplug, Zeng2021xvlm}, facilitating the learning of cross-modal representations from image-text pairs by aligning visual and linguistic modalities.  

Recent studies~\cite{Andonian2022RobustCR, Li2021AlignBF, li2022blip}  have found that SMCL-based models pre-trained on web-harvested data often suffer from data inefficiency since image captions frequently contain words that are unrelated to the image content or only capture partial information. One common strategy is increasing the scale of the training to alleviate the negative impacts of noisy data samples~\cite{clip, Jia2021ScalingUV, Yu2022CoCaCC}. A typical example is CLIP~\cite{clip}, which utilizes a massive dataset of 400 million image-text pairs obtained through web crawling. Though it demonstrated promising results in enhancing the cross-modal capabilities of models, scaling up datasets presents a challenge due to the high computational cost.  For example, CLIP requires an estimated 3584 GPU (V100) days for pertaining, a demand that is financially prohibitive under a constrained budget. Other researchers exploit soft labels~\cite{Li2021AlignBF,Andonian2022RobustCR} or regenerate image captions~\cite{li2022blip} to mitigate the impact of noisy data, yet with unsatisfactory performance improvement or substantial additional computation.

\begin{figure}[t]
\centering

\subfigure[VQA and Pre-training Time]{
\centering
\includegraphics[width=1.65in]{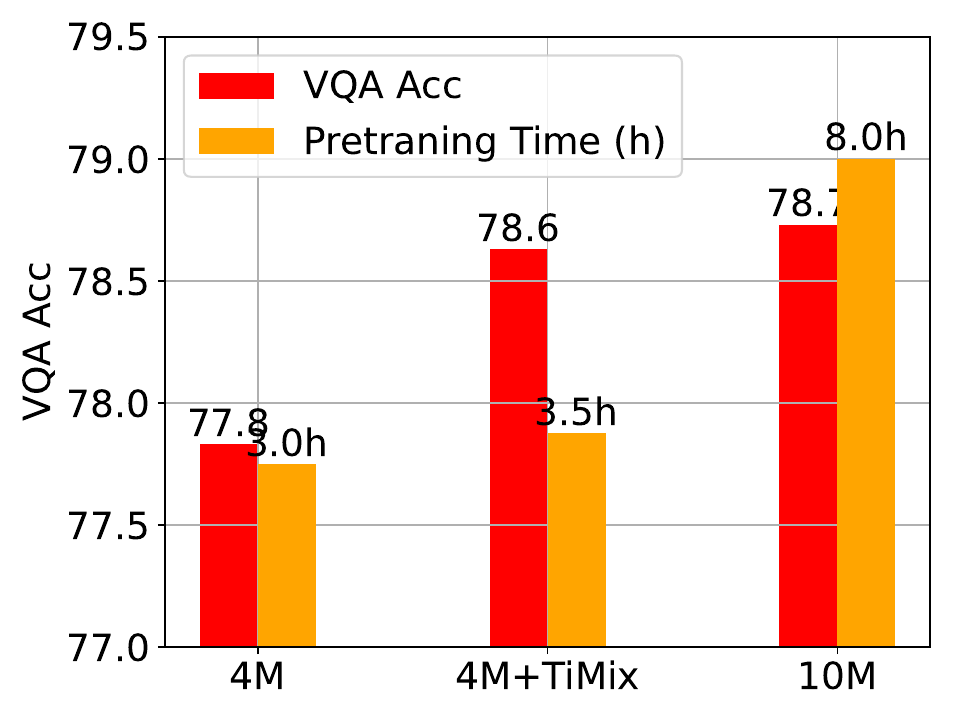}
}%
\subfigure[Contrastive Learning Loss ]{
\centering
\includegraphics[width=1.65in]{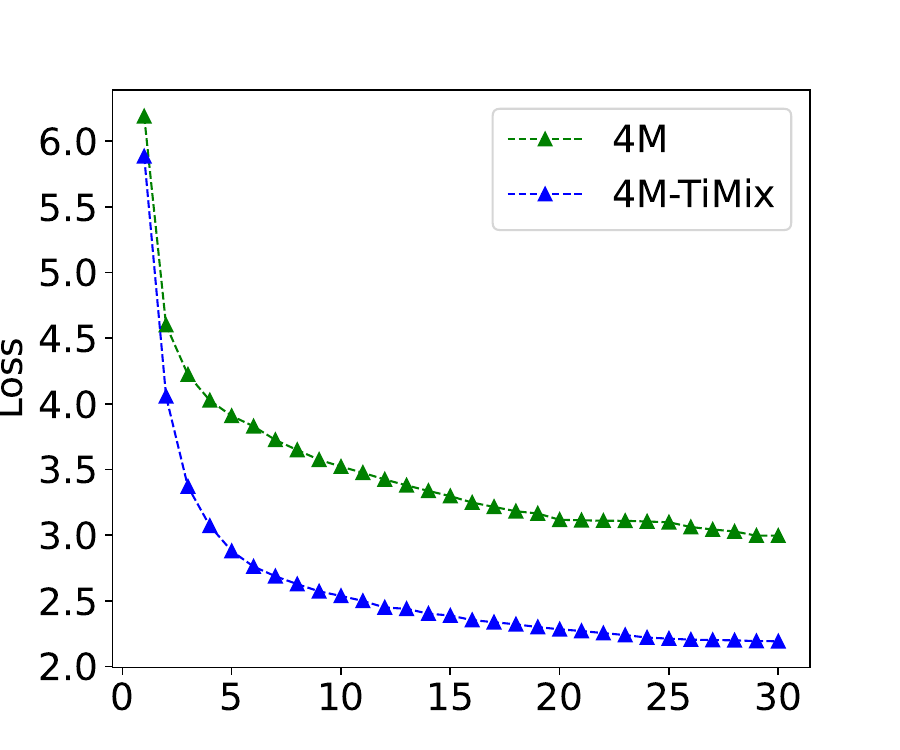}
}%
\caption{Subfigure (a) illustrates the Visual Question Answering results and pre-training time per epoch of the VLP model mPLUG~\cite{li2022mplug} which are pre-trained on with different data sizes on 8 $\times$ 80G A100. 4M+\modelname refers to training on 4M data with TiMix. Subfigure (b) illustrates the convergence curve of cross-modal contrastive learning, the x-axis is labeled as epoch.}
\label{fig:gpu training time}
\end{figure}
In this paper, we present a novel perspective of data mixing to tackle data inefficiency in VLP. We hypothesize that an image could exhibit multiple distinct views, each potentially associated with a different textual caption. These diverse textual descriptions align with specific views that capture various aspects of the image's semantic information. Building upon this hypothesis, we introduce Text-aware Image Mixing (\modelname), which adopts the CutMix~\cite{Yun2019CutMixRS} approach to create data samples for contrastive learning. Specifically, we design a patch-text alignment (PTA) pre-training task, allowing us to learn the matching degree between patches and captions. So we can mix two images guided by the relevance of their patches to their captions. Then the mixed samples are incorporated into contrastive learning to improve cross-modal representation and enhance data efficiency.

We theoretically analyze TiMix from a mutual information (MI) maximation perspective and find that mixed data samples implicitly provide a regularizer for the contrastive learning loss function. This regularizer keeps the model from overfitting to partially aligned image-text pairs during the contrastive learning process, thereby mitigating the negative impact of noisy data. Empirically, by incorporating TiMix into existing VLP models, we can observe consistent performance improvement on common vision-language downstream tasks, including Visual Question Answering (VQA), Cross-modal Retrieval, Natural Language for Visual Reasoning (NLVR) and Image Captioning, with small additional computational cost during training.

In summary, our contributions are: 
\begin{itemize}
\item We take the first step to introduce mix-based data samples into vision-language pre-training. With a novel patch-text alignment pre-training task, mixed images are created in a CutMix style based on the matching degree of their patches and captions, serving as high-quality data for cross-modal contrastive learning.

\item We theoretically prove that mixed data samples implicitly provide a regularizer for cross-modal contrastive learning, facilitating  mutual information optimization for potentially partially-aligned image-text pairs.

 \item Experimental findings illustrate that \modelname delivers robust performance, significantly enhancing data efficiency while maintaining cost-effectiveness during the pre-training phase. For example, as shown in Figure~\ref{fig:gpu training time}, \modelname achieves comparable downstream task performance by training on 40\% of the data in 43.8\% of the training time, compared to a recent robust VLP model mPLUG.




\end{itemize}




\section*{Related Work}

\subsection{Vision-Language pre-training}

Recent years have seen significant success for large-scale pre-trained vision-language models ~\cite{Tan2019LXMERTLC,Jiang2022TRIPS,Chen2020UNITERUI,Huang2020PixelBERTAI,Li2020OscarOA,Yu2021ERNIEViLKE,Li2021AlignBF,Wang2021SimVLMSV,li2022mplug, 2021VinVL, Kim2021ViLTVT,map,scl, ptp,beit3} in a variety of cross-modal tasks.  Current approaches to VLP can be broadly divided into two categories in terms of visual representation extraction. The first category is detector-based VLP methods \citep{Lu2019ViLBERTPT,Li2019VisualBERTAS,Tan2019LXMERTLC,Li2020OscarOA,Chen2020UNITERUI,Yu2021ERNIEViLKE,wang2020minivlm,gan2021playing, fang2021compressing}. These methods primarily adopt a two-step training pipeline: they extract visual features using a pre-trained object detector and then train the cross-modal pre-training model to align text and visual features. The main challenge for these methods is to balance effectiveness and efficiency. The second category consists of more recent CNN-based \citep{Huang2020PixelBERTAI,Xu2021E2EVLPEV} or ViTs-based \citep{Li2021AlignBF,Kim2021ViLTVT,Radford2021LearningTV,Wang2021VLMoUV} methods, especially patch-based ViT. These methods eliminate the need for a complex object detector in feature extraction, enabling end-to-end VL learning. Furthermore, Self-supervised Multi-modal Contrastive Learning (SMCL) has lately sparked significant advancements ~\cite{li2022blip,clip, yao2021filip, Li2021AlignBF, li2022mplug} by conducting cross-modal alignment. SMCL consists of image-to-text and text-to-image contrastive learning, e.g., with the InfoNCE~\cite{oord2018cpc} loss.

\subsection{Mixed Data Augmentation}
Mixup~\cite{Zhang2017mixupBE} is a widely used data augmentation technique in Computer Vision, which involves training by convexly combining the input image and its corresponding label. CutMix~\cite{Yun2019CutMixRS}, on the other hand, is a specific case of Mixup and can be seen as a pixel-level Mixup method that utilizes binary masks. Recent developments~\cite{Uddin2020SaliencyMixAS, Liu2022TokenMixRI,Walawalkar2020AttentiveCA} in Mixup methods have focused on effectively leveraging saliency information and performing mixing at the image feature level. The motivation behind saliency-based Mixup methods is to preserve salient regions when blending images, ensuring that sufficient information is retained and more natural feature representations are learned. SaliencyMix~\cite{Uddin2020SaliencyMixAS} employs various saliency detectors to directly extract salient regions from the images. Co-Mixup~\cite{Kim2021CoMixupSG} aims to maximize the gradient-based saliency while encouraging diversity in the mixed images' hypermodules. SuperMix~\cite{Dabouei2020SuperMixST} utilizes supervised signals to mix input images based on saliency. However, the majority of existing Mixup methods have mainly been applied in the CV. In contrast, we introduce Mixup techniques to the multimodal domain for the first time, resulting in notable advancements.

\section*{Method}

 Our approach exhibits similarities to the Cutmix-style method. Specifically, our method involves pairing two images within a batch, denoted as $I^x$ and $I^y$, along with their corresponding texts, $T^x$ and $T^y$. Assuming $I^y$ as the source image, we extract the region $\dot{R}^y$ in the source image $I^y$ that exhibits the highest text-relevant score with $T^y$. Subsequently, we replace the corresponding region $\dot{R}^x$ in the target image $I^x$, which possesses the lowest text-relevant score with $T^x$, with the extracted region $\dot{R}^y$. This process yields a mixed image denoted as $\hat{I}^{x,y}$. To determine the soft labels for the mixed image and text, we consider the proportion of the mixed and cropped regions. Additional details regarding this process are provided in the following section.

\begin{figure*}[t]
    \centering
    \includegraphics[width=0.99\textwidth]{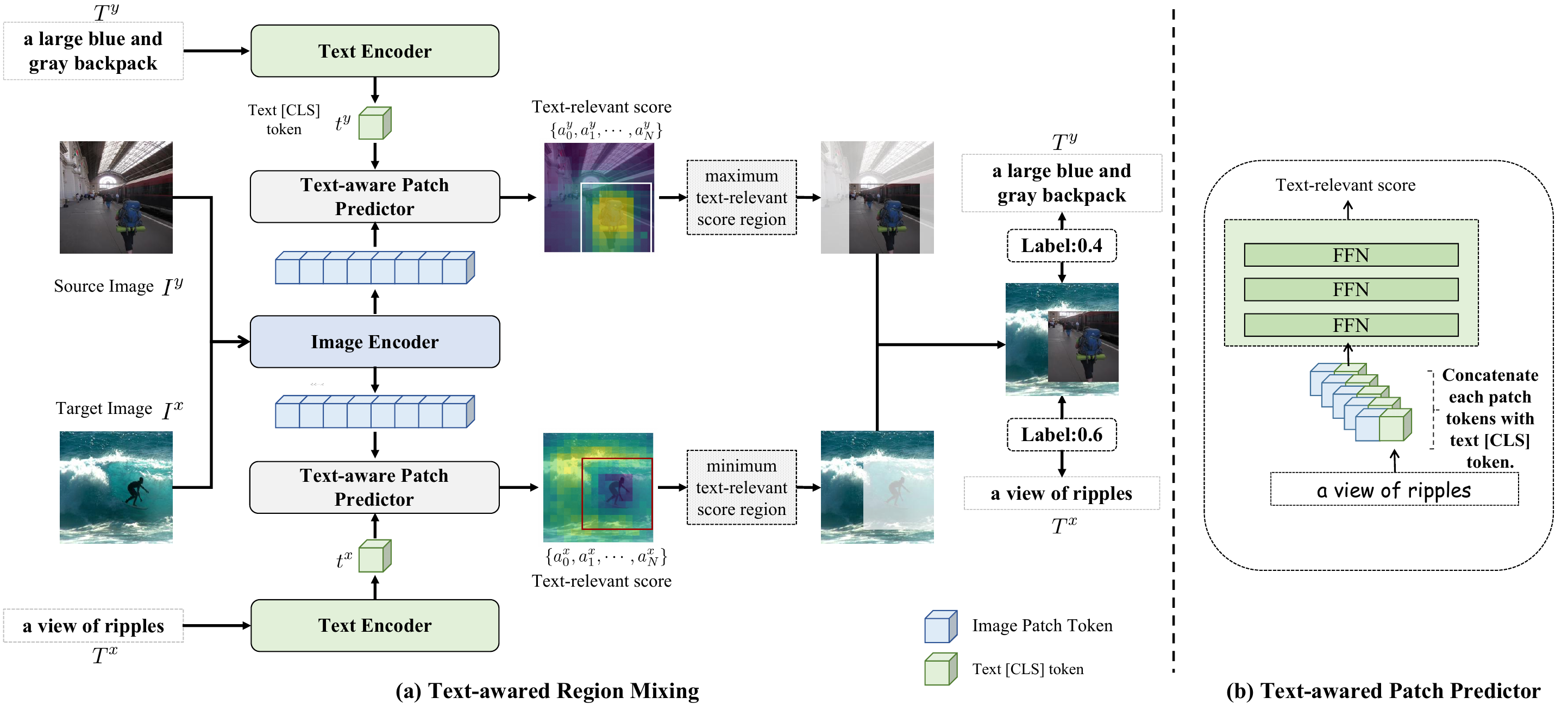}
    \caption{The subfigure (a) illustrates the process of TiMix, where two image-text pairs are utilized. Subfigure (b) depicts the architecture of the Text-aware patch predictor.}
    \label{fig:pipline}
    \vspace{-3ex}
\end{figure*}

\subsection{Text-aware Region Mixing}

As depicted in Figure~\ref{fig:pipline}, considering two pairs of image-text denoted as $I^x, T^x$ and $I^y, T^y$ respectively, let us designate $I^x$ as the target image and $I^y$ as the source image. Initially, suppose the image size is $H \times W$,  the target image $I^x$ and the source image $I^y$ are divided into distinct, non-overlapping patches of size $P \times P$, resulting in a total of   $\frac{H}{P} \times \frac{W}{P}$ patches. Subsequently, these patches are fed into the visual backbone to extract the sequential representation of visual features. Consequently, the extracted representations of $I^x$ and $I^y$ assume the form of $\{v^x_{cls}, v^x_1, \dots, v^x_{N}\}$ and $\{v^y_{cls}, v^y_1, \dots, v^y_{N}\}$ respectively, where $N =\frac{H}{P} \times \frac{W}{P}$. Then, both textual captions $T^x$ and $T^y$ are fed to the text backbone and the [CLS] tokens symbolized as $t^x \in R^D$ and $t^y \in R^D$ are employed to encapsulate the comprehensive context of the text, D is the dimension of text representation. Next, each visual token in the sequence $\{v^x_{cls}, v^x_1, \dots, v^x_{N}\}$ is concatenated with $t^x$ and passed through the Text-aware Patch Predictor (TPP) to compute the text-relevant scores for the patches, denoted as $A_x = \{a^x_1,\dots,a^x_N\}$. Similarly, for the sequence $\{v^y_{cls}, v^y_1, \dots, v^y_{N}\}$, each visual token is concatenated with $t^y$, and fed to the text-relevant scores to calculate the text-relevant scores of patches $A_y = \{a^y_1,\dots,a^y_N\}$, where $A_x, A_y \in R^{N}$. We rearrange the shape of $A_x, A_y$ to $\frac{H}{P} \times \frac{W}{P}$.

We aim to extract a region $\hat{R}^y$ from the source image $I^y$ and merge it with the target image $I^x$ to create a mixed image $\hat{I}^{x,y}$. For this purpose, we introduce a side ratio $\gamma$, which is sampled from a uniform distribution ranging from 0.25 to 0.75. This ratio helps determine the total number of patches, which is given by $\lfloor\gamma\frac{H}{P} \rfloor \times \lfloor\gamma\frac{W}{P}\rfloor$, that will be obtained from the cropped region. Given a paired text, to calculate the overall text-relevant score of each region with the text, We employ a 2D convolution operation on $A_x$ and $A_y$ with a kernel size of $\lfloor\gamma\frac{H}{P} \rfloor \times \lfloor\gamma\frac{W}{P}\rfloor$ and a stride of 1 to iterate through all the regions. The center indices ( denoted as a, b) of the regions should satisfy the following conditions:

\small
\begin{align}
    & a^x,b^x =  \mathop{\mathrm{argmin}}\limits_{a,b} \sum_{p,q} A^x_{a+p-\lfloor\frac{h}{2}\rfloor, b+q-\lfloor\frac{w}{2}\rfloor}\\
    & a^y,b^y =  \mathop{\mathrm{argmax}}\limits_{a,b} \sum_{p,q} A^y_{a+p-\lfloor\frac{h}{2}\rfloor, b+q-\lfloor\frac{w}{2}\rfloor}
\end{align}
\normalsize

Where $h=\lfloor\gamma\frac{H}{P} \rfloor, w=\lfloor\gamma\frac{W}{P}\rfloor, p\in \{0,1,\dots,h-1\}. q \in \{0,1,\dots,w-1\}$. Then, we obtain the new mixed image sample $\hat{I}^{x,y}$ as follows:

\small
\begin{align}
    &\hat{I}^{x,y} = I^x \\
    &\hat{I}^{x,y}_{a^x+p-\lfloor\frac{h}{2}\rfloor, b^x+q-\lfloor\frac{w}{2}\rfloor} = I^y_{a^y+p-\lfloor\frac{h}{2}\rfloor, b^y+q-\lfloor\frac{w}{2}\rfloor}
\end{align}
\normalsize
Then the soft label of the mixed image $\hat{I}^{x,y}$ to text $t^y$ is calculated as follow:

\small
\begin{align}
    &\mathbf{s}^y = \frac{hwP^2}{HW}
\end{align}
\normalsize

The soft label of the mixed image $\hat{I}^{x,y}$ to text $t^x$ is calculated as $\mathbf{s}^x = 1-\mathbf{s}^y$.


\subsection{Learning the Text-relevant Score of Patch}
\label{subsec:PTA}
The key component of \modelname is the text-aware patch predictor which needs to predict text-relevant scores between the image patches and input text. As shown in Figure~\ref{fig:pipline}, the patch predictor is a Multi-Layer Perceptron (MLP) that contains three linear layers and is used to predict the alignment score between patches and the input text. 

As the lack of fine-grained patch-text labels to train the text-aware patch predictor, in this sub-section, we propose to convert object-level signals into patch-level ones and introduce a novel pre-training task named Patch Text Alignment which facilitates the patch predictor training and drives our model to learn the fine-grained patch-text alignment. For object objection and visual grounding datasets like COCO\cite{Lin2014MicrosoftCC} and VG\cite{Krishna2016VisualGC}, the object and region generally be paired with a class label or text description. Therefore, we can transfer every object class label to a text description based on a text template such as "This is a [class label]". Thus, for each (object/region) bounding box in an image, we can construct a text description for it. Then, we transform the bounding box annotations to the patch-level labels by following this rule:  Given an image and a bounding box annotation, if there is an overlap between an image patch and a bounding box, it will be assigned with label 1, otherwise, it will be assigned with label 0.
For different text descriptions and bounding boxes, the labels of the patch are different. In this way, we can generate fine-grained patch-text labels which can be served as the supervisory signal to pre-train our model.

After that, in each step of pre-training, we randomly sample a mini-batch of images from the object detection/visual grounding datasets. For each image, we randomly select an object/region bounding box and translate the bounding box annotation to the image patch label sequence following the transformation rule we mentioned before. Then, we feed the batch of text descriptions of the bounding boxes and the images together to our VLP model.  We hope the text-aware patch predictor can detect all patches which have overlap with the bounding box with the guidance of the bounding box text description.  Supposing the text-aware patch predictor has predicted the text-relevant scores between image patches and text, we calculate the binary cross entropy loss between the text-relevant scores and patch labels as:
    \begin{equation}
        \mathcal{L}_{PTA} = \frac{1}{e}\sum_{i=1}^{e} Y_{i} log\left(a_i\right) + \left(1-Y_{i}\right)log\left(1-a_i\right) 
    \end{equation}
where $a_i$ is the text-relevant score between $i_{th}$ patch in the image and the input text, $Y_i$ is the patch label of $i_{th}$ patch. 

\subsection{Contrastive Learning Based on \modelname }
\begin{figure}
  \centering
    \includegraphics[width=0.45\textwidth]{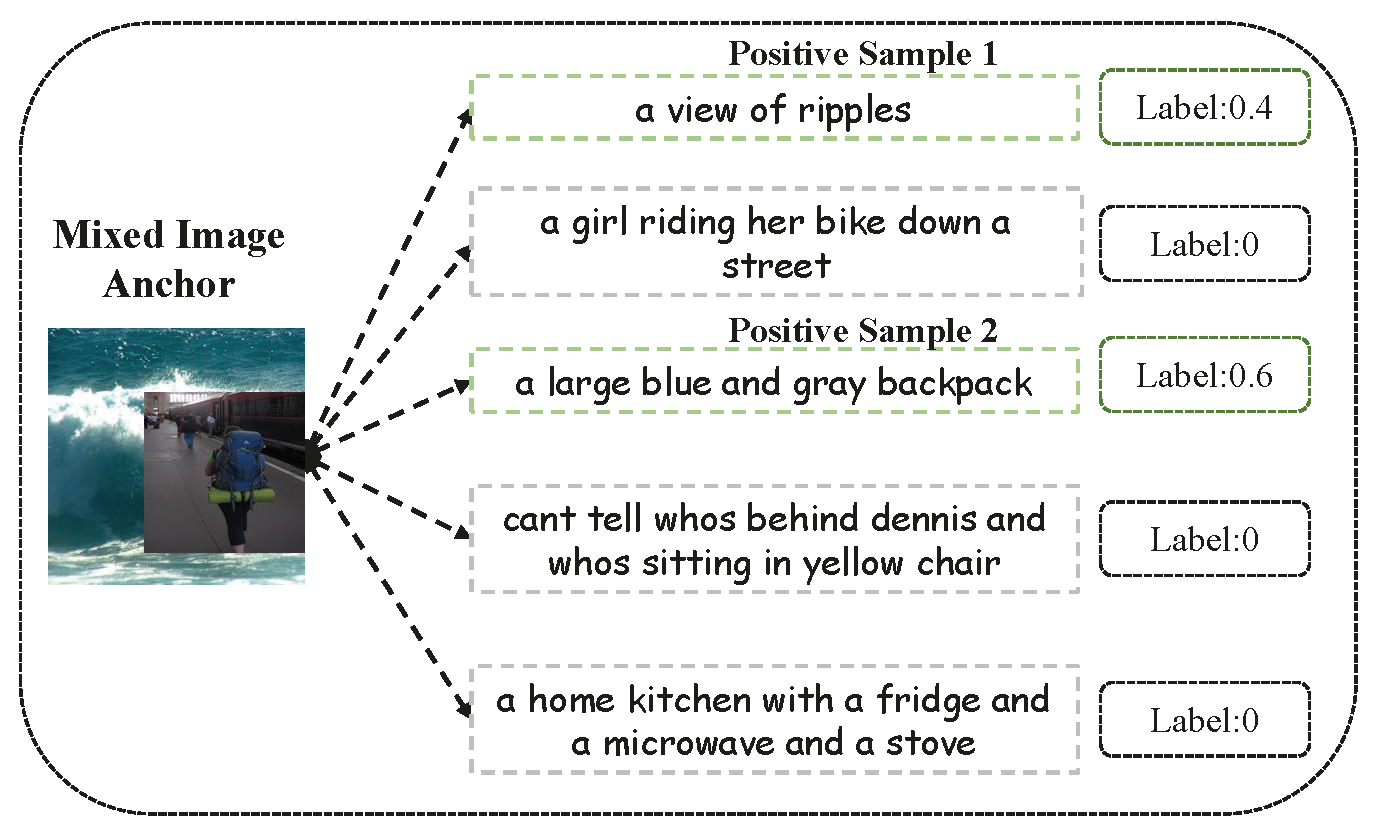}
    \caption{An example of TiMix in image-to-text contrastive learning. The text within the green box represents the positive samples, and the text within the gray box represents the negative samples.}
    \label{fig:fig3}

\end{figure}
In this subsection, we will introduce how to apply TiMix to Vision Language Pretraining (VLP) with unsupervised cross-modal contrastive learning.

Let us consider a random sampling of N image-text pairs to compose a minibatch. Within this set, we form pairs randomly which results in N groups, each N consisting of two image-text pairs. Let us denote the two pair as $I^x, T^x$ and $I^y, T^y$. Applying the aforementioned method, we use $I^x$ as the target image and $I^y$ as the source image to generate the mixed image $\hat{I}^{x,y}$. Similarly, by swapping the roles, with $I^y$ as the target image and $I^x$ as the source image, we obtain the mixed image $\hat{I}^{y,x}$. By repeating this process, we acquire N mixed images. As shown in Figure~\ref{fig:fig3}, we select a mixed image sample $\hat{I}^{x,y}$ as the anchor. The two texts $T^x$ and $T^y$ associated with the source image and target image are regarded as positive samples. Following the aforementioned procedure, we compute the soft labels of the anchor image to these two positive samples. The remaining texts are considered negative samples. Suppose the extracted global vision representation of the mixed image is denoted as $\hat{v}^{x,y}$ and there are two positive text samples, and the global representations of them are denoted as $t^x$ and $t^y$. Then, the image-to-text contrastive loss based on \modelname can be formulated as follow:

\begin{align}
 & \mathcal{L}_{\modelname}^v= \\ \nonumber
& -\sum\limits_{i = 1:\mathcal{N}}\frac{1}{\mathcal{N}}\Bigg[ \mathbf{s}^x\mathop{log} 
                                \Bigg[ 
                                \frac{
                                    f\left( \hat{v}^{x,y}_i, t^x_i \right)
                                }
                                {
                                    f\left(  \hat{v}^{x,y}_i, t^x_i\right) + \sum\limits_{t_k \neq t^x_i} 
                                    f\left( \hat{v}^{x,y}_i , t_k\right)
                                 }
                                \Bigg] \\  \nonumber
& +\mathbf{s}^y 
                                \mathop{log} 
                                \Bigg[ 
                                \frac{
                                     f\left( \hat{v}^{x,y}_i, t^y_i \right)
                                }
                                {
                                    f\left(  \hat{v}^{x,y}_i, t^y\right) + \sum\limits_{t_k \neq t^y_i} 
                                    f\left(  \hat{v}^{x,y}_i, t_k\right)
                                 }
                                \Bigg]\Bigg]
\label{eq:contras1}
\end{align}

where $f\left(\hat{v}^{x,y}, t^x\right)$ measures the distance between $\hat{v}^{x,y}$ and $t^x$ in a semantic space, $\mathcal{N}$ represents the number of batch size. For text-to-image contrastive learning, within a batch, a specific text is paired with its corresponding image, which is used as both the source image and the target image to generate two mixed images. These two mixed images are considered positive samples, while the other mixed samples in the batch are treated as negative samples. The label assigned to the text anchor with respect to these two positive image samples is the same as the labels used in image-to-text contrastive learning.

\section*{A Mutual Information Maximization Perspective}

In this section, we provide evidence and explanations for our method from the perspective of maximizing mutual information. Following the definition in \cite{oord2018cpc} in the context of image-to-text contrastive learning, the similarity function $f\left( v_i, t_i\right)$ in Equation \ref{eq:contras1} can be utilized to model the density ratio, which preserves the mutual information between the image $v_i$ and the text $t_i$ and we rewrite the $f\left( v_i, t_i\right)$ to $ \frac{\mathop{P}\left( t_i | v_i \right)}{\mathop{P}\left( t_i\right)} $.


Then, given a batch of unmixed image-text pairs, the vanilla contrastive learning loss $ \mathcal{L}^v$ satisfies the following inequality:
\begin{align}
&\mathcal{L}^v= -\mathop{E}\limits_{t} \mathop{log} 
                                \left[
                                 \frac{ 
                                   \frac{\mathop{P}\left( t_i \middle| v_i \right)}
                                         {\mathop{P}\left( t_i \right)}
                                }
                                { 
                                    \frac{\mathop{P}\left( t_i  \middle|v_i \right)}
                                         {\mathop{P}\left( t_i \right)} 
                                      +
                                    \sum\limits_{k \neq i }
                                        \frac{\mathop{P}\left( t_k \middle| v_i\right)}
                                             {\mathop{P}\left( t_k \right)} 
                                }
                                \right] \\
        &\geq -\mathop{I}(t_i, v_i) + log\left(N\right) \label{eq:vanilla_eq}
\end{align}
where $\mathop{I}(t_i, v_i)$ denotes the mutual information between $t_i$ and $v_i$.
The detailed proof can be found in Appendix~\ref{proof A}. 
Based on inequality~\ref{eq:vanilla_eq}, we can get the lower bound of $\mathop{I}(t_i, v_i)$ as:
\begin{equation}
   \mathop{I}(t_i, v_i) \geq log\left(N\right)- \mathcal{L}^v \label{eq:eqIM1}
\end{equation}
With a similar derivation, we can get another inequality about the image-to-text contrastive learning loss in \modelname as follows:
\begin{align}
&\mathcal{L}_{\modelname}^v= -\mathop{E}\limits_{t} \mathbf{s}^x  \mathop{log} 
                                \left[
                                 \frac{ 
                                   \frac{\mathop{P}\left( t^x \middle| \hat{v}^{x,y} \right)}
                                         {\mathop{P}\left( t^x\right)}
                                }
                                { 
                                    \frac{\mathop{P}\left( t^x \middle| \hat{v}^{x,y} \right)}
                                         {\mathop{P}\left( t^x\right)} 
                                      +
                                    \sum\limits_{k \neq i }
                                        \frac{\mathop{P}\left( t_k \middle| \hat{v}^{x,y}\right)}
                                             {\mathop{P}\left( t_k \right)} 
                                }
                                \right]\nonumber  \\
                            &    -\mathop{E}\limits_{t}
                                \mathbf{s}^y
                                \left[
                                  \frac{ 
                                   \frac{\mathop{P}\left( t^y \middle| \hat{v}^{x,y} \right)}
                                         {\mathop{P}\left( t^y\right)}
                                }
                                { 
                                    \frac{\mathop{P}\left( t^y \middle|\hat{v}^{x,y} \right)}
                                         {\mathop{P}\left( t^y\right)} +
                                     \sum\limits_{k \neq j }
                                        \frac{\mathop{P}\left( t_k \middle| \hat{v}^{x,y}\right)}
                                             {\mathop{P}\left( t_k \right)} 
                                } \right]   \\
        & \geq -\mathbf{s}^x\mathop{I}(t^x, \hat{v}^{x,y}) - \mathbf{s}^y\mathop{I}(t^y, \hat{v}^{x,y}) + (\mathbf{s}^x+ \mathbf{s}^y)log\left(N\right)\label{sec4:eq3}
\end{align}
\normalsize
 where the $\mathop{I}(t^x, \hat{v}^{x,y})$ is the mutual information between $t^x$ and $\hat{v}^{x,y}$. 
 
 Noted that given a mixed image $\hat{v}^{x,y}$, there is a target image $v^x$ paired with the text $t^x$ and a source image $v^y$ paired with the text $t^y$. Suppose $v_{\textbf{r}}^x$ denotes the region in $v^x$ that has the maximum text-relevant score with $t^x$ and $v_{\textbf{r}}^y$ denotes the region in $v^y$ that has the maximum text-relevant score with $t^y$, the mixed image $\hat{v}^{x,y}$ can be seen as the combination of $v_{\textbf{r}}^x$ and $v_{\textbf{r}}^y$. Then, based on the chain rule of mutual information, we can get the lower bound of $\mathbf{s}^x \mathop{I}(t^x, v_{\textbf{r}}^x) +  \mathbf{s}^y\mathop{I}(t^y, v_{\textbf{r}}^y)$ as:
 
\small
\begin{align}
     &\mathbf{s}^x \mathop{I}(t^x, v_{\textbf{r}}^x) +  \mathbf{s}^y\mathop{I}(t^y, v_{\textbf{r}}^y) \nonumber  \\
     & \geq   log\left(N \right) - (\mathcal{L}_{\modelname}^v + \mathbf{s}^x\mathop{I}(t^x, v_{\textbf{r}}^y)+\mathbf{s}^y \mathop{I}(t^y, v_{\textbf{r}}^x) ) 
     \label{eq:eqIM2}
\end{align}
\normalsize
The details of this derivation can be found in appendix \ref{proof B}. Combining inequality ~\ref{eq:eqIM1} and ~\ref{eq:eqIM2} provides us with inspirational findings. It has been easily seen (and widely known) that traditional InfoNCE loss tries to maximize the lower bound of mutual information between $t_i$ and $v_i$. Similarly, in the scenario of \modelname, the lower bound of $\mathop{I}(t^x, v_{\textbf{r}}^x)$ and $\mathop{I}(t^y, v_{\textbf{r}}^y)$ should be maximized given ideal clean data. Note that on the right side of  inequality \ref{eq:eqIM2}, $\mathcal{L}_{\modelname}^v$ is attained with two items $\mathop{I}(t^x, v_{\textbf{r}}^y)$ and $\mathop{I}(t^y, v_{\textbf{r}}^x)$. These two elements effectively act as implicit regularizers, preventing $\mathcal{L}_{\modelname}^v$ from becoming excessively optimized. This mitigates the risk of over-maximizing the mutual information terms $\mathop{I}(t^x, v_{\textbf{r}}^x)$ and $\mathop{I}(t^y, v_{\textbf{r}}^y)$, thereby making the model more robust in the context of inaccurately or partially aligned image-text pairs and hence leading to an improvement of data efficiency.

\section*{Experiment}
Following the previous works ~\cite{Li2021AlignBF} and \cite{li2022mplug}, we use the same pre-training dataset with 14M images with texts, which includes two in-domain datasets (MS COCO ~\cite{Lin2014MicrosoftCC} and Visual Genome ~\cite{Krishna2016VisualGC}), and three web out-domain datasets (Conceptual Captions ~\cite{cc}, Conceptual 12M ~\cite{changpinyo2021conceptual}, SBU Captions ~\cite{Ordonez2011Im2TextDI}).  Please refer to Appendix ~\ref{sup:setting} to see more detail about the pre-training dataset and pre-training setting. 

\subsection{Overall Performance}
We evaluate \modelname with two well-known VLP models ALBEF and mPLUG (denoted as ALBEF-\modelname and mPLUG-\modelname) on four vision-language downstream tasks: visual question answering (VQA2.0 \cite{Agrawal2015VQAVQ}), natural language for visual reasoning (NLVR2 \cite{NLVR2}), image-text retrieval(Flickr30K \cite{Plummer2015Flickr30kEC}, COCO \cite{Lin2014MicrosoftCC}), image captioning(COCO Caption \cite{Lin2014MicrosoftCC}). Our baselines cover 16 VLP models, detailed in Appendix~\ref{sup:comparison models}(In our experiments, we only re-implement the base version of ALBEF \cite{Li2021AlignBF} and mPLUG \cite{li2022mplug}.). We will first analyze their overall performances on these tasks. The fine-tuning hyper-parameters and the details of downstream tasks are described in Appendix \ref{sup:downsteam}.
\begin{table}[t]
\centering
\small
\setlength\tabcolsep{5pt}
\begin{tabular}{l|c|cc|cc}
\toprule[1.5pt]
\multirow{2}{*}{model} & \multirow{2}{*}{\begin{tabular}[c]{@{}c@{}} Pre-train\\ Data \end{tabular}} & \multicolumn{2}{c|}{VQA} & \multicolumn{2}{c}{NLVR2} \\
                       &                                                                                   & dev   & std   & Dev        & Test-P       \\ \hline
VisualBERT  & 180K & 70.80 & 71.00&67.40 & 67.00\\
LXMERT & 180K & 72.42 & 72.54 & 74.90 & 74.50\\
ViLBERT  & 3.3M  & 70.63 & 70.92 & - & - \\
E2E-VLP& 4M & 73.25  & 73.67 &    77.25 & 77.96      \\
UNITER & 4M & 73.82 & 74.02 & 79.12 & 79.98 \\
METER & 4M & 77.68 & 77.64  & 82.33 & 83.05 \\
ViLT  & 4M  & 71.26 & 71.29  & 75.18  & 76.2 \\
VLMo  & 4M & 76.64 & 76.89  & 82.77 & 83.34 \\
OSCAR  & 6.5M & 73.16 & 73.44 & 78.07 & 78.36 \\
VinVL & 5.65M & 76.52 & 76.60 & 82.67 & 83.98 \\
XVLM  & 14M & 78.22& 78.37 &84.41 & 84.76\\ 
BLIP  & 129M & 78.25 & 78.32 & 82.48 & 83.08\\
MAP & 4M &78.03 &- & 83.30 & 83.48  \\
SCL & 4M & 78.72  & 78.78  & 83.63 & 84.27 \\
SimVLM  & 1.8B & 77.87 & 78.14  & 81.72 & 81.77  \\   \hline
ALBEF  & 4M  &  74.54 & 74.70 & 80.24 & 80.50 \\
\rowcolor{gray!20}ALBEF -\modelname & 4M & 75.92 & 76.23 & 82.42 & 83.03\\  \hline
ALBEF  & 14M  & 75.84 &76.04 & 82.55& 83.14 \\
\rowcolor{gray!20} ALBEF -\modelname & 14M & 76.82 &77.11 & 83.44& 83.47 \\  \hline
mPLUG  & 4M & 77.83 & 77.98  & 82.66 & 82.92  \\
\rowcolor{gray!20} mPLUG -\modelname& 4M & 78.63  & 78.85 & 84.12 & 84.23\\   \hline
mPLUG  & 14M & 79.65  & 79.22 & 83.43 & 84.21 \\
\rowcolor{gray!20} mPLUG -\modelname & 14M & \textbf{80.83}  & \textbf{81.53} & \textbf{84.77} & \textbf{85.22}\\
\bottomrule[1.5pt]
\end{tabular}
\caption{Evaluation results on VQA2.0 and NLVR$^2$. More details about comparison models are in Appendix~\ref{sup:comparison models}. } 
\setlength \tabcolsep{1.0pt}
\label{table:vqa_caption}
 
\end{table}
\begin{table*}[t]

\setlength\tabcolsep{5.5pt}
\centering
\small
\begin{tabular}{l|c|cccccc|cccccc}
\toprule[1.5pt]

\multicolumn{1}{c|}{\multirow{2}{*}{Models}}      &
\multicolumn{1}{c|}{\# Pre-train} &
\multicolumn{6}{c|}{MSCOCO (5K test set)} & \multicolumn{6}{c}{Flickr30K (1K test set)} \\
      &  data & \multicolumn{3}{c}{TR} & \multicolumn{3}{c|}{IR} & \multicolumn{3}{c}{TR} & \multicolumn{3}{c}{IR}          \\
\midrule
&&R@1&R@5&R@10&R@1&R@5&R@10&R@1&R@5&R@10&R@1&R@5&R@10 \\ \hline
E2E-VLP & 4M     &-& -&-&-&-&- & 86.2 &97.5 &98.92&73.6 & 92.4 &96.0 \\
OSCAR  & 4M  & 70.0&91.1&95.5&54.0&80.8&88.5&-& -&-&-&-&-   \\
VinVL  & 5.65M & 75.4 & 92.9 & 96.2 &58.8 &83.5 &90.3 &-& -&-&-&-&-   \\
ViLBert & 3.3M  & - &- &- &- &- &-s& -& -&-&58.2 & 84.9 & 91.5  \\
UNITER & 4M     & 65.7&88.6&93.8&52.9&79.9&88.0&87.3& 98.0&99.2&75.6&94.1&96.8  \\
METER  & 4M & 76.2 & 93.2 & 96.8& 57.1 & 82.7 & 90.1 &94.3 & 99.6 & 99.9  &82.2  &96.3 & 98.4  \\
VLMo  & 4M & 78.2& 94.4& 97.4& 60.6& 84.4& 91.0& 95.3& 99.9& 100.0& 84.5& 97.3& 98.6 \\
BLIP  & 129M &  \textbf{82.4} & 95.4&97.9&65.1&86.3&91.8&97.4& 99.8&99.9&87.6&97.7&99.0                   \\
SCL & 4M& 77.7 & 94.1 &97.4& 60.1& 84.6& 91.5& 95.9& 99.8& 100.0 &84.6& 97.4 &98.9 \\
MAP & 4M& 79.3 & 94.8& 97.6& 60.9 &86.2& 93.1  &94.9& 99.5 &99.8& 83.8& 97.2 &98.7 \\
ViLT & 4M  & 61.5 & 86.3 &92.7& 42.7& 72.9&83.1&83.5& 96.7 &98.6 &64.4 &88.7&93.8\\
 \hline
ALBEF  & 4M & 73.1 & 91.4 & 96.0 & 56.8&  81.5&  89.2&  94.3 & 99.4 &99.8& 82.8& 96.7 & 98.4   \\
\rowcolor{gray!20}ALBEF -\modelname & 4M & 76.4 & 93.7 & 96.6 & 60.4 & 83.2 & 90.3&95.1& 99.8 &100.0 & 84.2 & 97.3 & 98.6       \\\hline

ALBEF  & 14M & 77.6&94.3&97.2&60.7&84.3&90.5&95.9& 99.8&100.0&85.6&97.5&98.9              \\
\rowcolor{gray!20} ALBEF -\modelname & 14M  &  78.8  & 95.2 &  97.6 & 61.3 & 85.2 & 91.0 & 96.7 & 99.8&100.0&86.4 & 97.2 & 99.0   \\\hline

mPLUG & 4M  & 78.8 & 94.1 & 96.4 & 61.2 & 85.2 & 90.6 & 95.9 & 99.8 &100.0& 85.7 & 96.8& 98.6   \\
\rowcolor{gray!20} mPLUG -\modelname & 4M  &  80.5 & 95.3 & 97.2 & 63.3 & 85.4  & 91.5 & 96.8 & 99.8&100.0&86.2 & 97.6 & 98.8   \\\hline

mPLUG & 14M  & 80.6 & 94.8 & 97.1 & 63.9 & 85.5 &91.2 & 96.5& 99.8&100.0& 86.3& 97.2& 98.9   \\
\rowcolor{gray!20} mPLUG -\modelname & 14M  &82.3 &\textbf{95.8}&\textbf{98.0}&\textbf{65.2}&\textbf{87.0}&\textbf{92.1}&\textbf{97.2}& \textbf{99.8}&\textbf{100.0}&\textbf{87.9}&\textbf{97.8}&\textbf{99.0}   \\
\bottomrule[1.5pt]
\end{tabular}      
\caption{Evaluation results of image-text retrieval on Flickr30K~\citep{Plummer2015Flickr30kEC} and COCO datasets~\citep{Lin2014MicrosoftCC}.}
\label{table:retrieval}
\end{table*}
\begin{table*}[t]
\centering
\small
\setlength\tabcolsep{13.2pt}
\begin{tabular}{l|c|cccccccc}
\toprule[1.5pt]
\multirow{3}{*}{Models} & \multirow{3}{*}{\begin{tabular}[c]{@{}c@{}}Pre-train \\ Data\end{tabular}} & \multicolumn{8}{c}{COCO Caption}                                                        \\
                        &                                                                              & \multicolumn{4}{c}{Cross-entropy Optimization} & \multicolumn{4}{c}{CIDEr Optimization} \\
                        &                                                                              & B@4         & M         & C         & S        & B@4       & M       & C       & S \\  \midrule     
E2E-VLP & 4M  & 36.2 &-&117.3&-&  - & - & - & - \\
OSCAR & 6.5M & 36.5 & 30.3 & 123.7 & 23.1 & 40.5 & 29.7 & 137.6 & 22.8  \\
VinVL  & 5.65M  & 38.5 & 30.4 & 130.8 & 23.4 & 41.0 & 31.1 & 140.9 & 25.2  \\
BLIP & 14M & 38.6 & - & 129.7 & - & - & - & - & - \\
LEMON  & 200M & 40.6 & 30.4 & 135.7 & 23.5 & 42.3 & 31.2 & 144.3 & 25.3 \\
UFO & 4M & 38.7 & 30.0 & 131.2 & 23.3 & - & - & - & -  \\
SimVLM & 1.8B & 39.0 & \textbf{32.9} & 134.8 & 24.0 & - & - & - & -  \\  \hline
mPLUG  & 4M  & 39.5 & 30.9 & 132.6 & 23.2 & 41.4 & 31.0 & 140.7 & 25.3   \\
\rowcolor{gray!20} mPLUG-\modelname & 4M & 40.7 & 31.2 & 134.8 & 23.9 & 41.5 & 30.9 &  141.2 & 25.4 \\\hline 
mPLUG  & 14M &  41.4 & 31.0 & 136.8 & 23.6 & 44.8 & 31.3 & 148.2 & 25.8   \\
\rowcolor{gray!20} mPLUG-\modelname & 14M & \textbf{42.2} & 31.4 & \textbf{138.3} & \textbf{24.1} & \textbf{45.2} &  \textbf{31.8} & \textbf{151.1} & \textbf{26.0}  \\
\bottomrule[1.5pt]
\end{tabular}
\caption{Evaluation Results on image captioning on COCO Karpathy test split \citep{karpathy2015deep}. B@4: BLEU@4, M: METEOR, C: CIDEr, S: SPICE. }
\label{table:captioning}
 
\end{table*}
\begin{table}[t]
\small
\centering
\setlength{\tabcolsep}{17pt}{
\begin{tabular}{@{}cccccc@{}}
\toprule[1.5pt]
\multicolumn{1}{c}{\multirow{2}{*}{Model}} & \multicolumn{3}{c}{RefCOCO+} \\
\multicolumn{1}{c}{}          & val   & testA   & testB          \\ \midrule
UNITER  & 75.90    & 81.45   & 66.70         \\
ViLBERT  & 72.34&  78.52 &   62.61             \\
VILLA & 76.17    & 81.54   & 66.84            \\
MDETR   & 79.52    & 84.09   & 70.62      \\
UNICORN & 80.30    & 85.05   & 71.88    \\ 
mPLUG & 80.07  &   85.21 &  71.03       \\
mPLUG-\modelname & \textbf{82.11}  &   \textbf{87.03} &  \textbf{74.43}    \\ 
\bottomrule[1.5pt]
\end{tabular}}
\caption{Evaluation results of visual grounding on ReferCOCO+. We use the accuracy of IOU 0.5 on visual grounding (a prediction is right if the IoU between the grounding-truth box and the predicted bounding box is larger than 0.5). Both mPLUG and mPLUG is pretrained on 4M datas.}
\label{tab:visual_grounding}
\end{table}

\subsubsection{Visual Question Answering}

The VQA task requires the model to answer natural language questions given an image. Following the approach of \cite{Li2021AlignBF}, we treat VQA as an answer generation problem. We evaluated our models by submitting the results to the evaluation server \footnote{https://eval.ai/web/challenges/challenge-page/830/leaderboard} and report the test-dev and test-std scores in Table~\ref{table:vqa_caption}. The VLP models equipped with \modelname demonstrate improved performance on the VQA task compared to the models without \modelname. These results highlight the significant improvements achieved by \modelname. Additionally, our mPLUG-TiMix model trained on 14M data outperforms other baseline models which provides further evidence of the effectiveness of our method.

\subsubsection{Natural Language for Visual Reasoning}

The NLVR2 \citep{NLVR2} task requires the model to predict whether a sentence accurately describes a pair of images, which is a binary classification task. For ALBEF-\modelname and mPLUG-\modelname, we follow \citep{Li2021AlignBF} and use two cross-attention layers to process the two input images; their outputs are merged and fed into a Feed Forward Network (FFN). As shown in Table~\ref{table:vqa_caption}, pre-trained with 14M, mPLUG-\modelname can obtain competitive  performances to the SOTA models.

\subsubsection{Image-Text Retrieval}

We conduct experiments for both image-to-text retrieval (TR) and text-to-image retrieval (IR) on MSCOCO \citep{Lin2014MicrosoftCC} and Flickr30K \citep{Plummer2015Flickr30kEC} datasets. As shown in Table \ref{table:retrieval},  pre-trained with 14M images, mPLUG-\modelname outperforms all existing methods on both datasets which even achieves better performance than BLIP with 129M. In addition, all models equipped with TiMix show significant improvements compared to their counterparts without TiMix.


\subsubsection{Image Captioning} 
 
 Following~\cite{li2022mplug}, we fine-tune mPLUG/ mPLUG-\modelname with cross-entropy loss and then with CIDEr optimization for an additional 5 epochs. Our experiments, as shown in Table~\ref{table:captioning}, unequivocally illustrate the superiority of mPLUG-\modelname over mPLUG alone. Notably, mPLUG-\modelname achieves performance levels that are comparable to those of SOTA models.

\subsubsection{Visual Grounding}

Following the setting of mPLUG~\citep{li2022mplug}, we also evaluate mPLUG-\modelname on the visual grounding task, which requires models to localize the referred object in the image based on a given text query. Instead of directly regressing the bounding boxes, we concatenate visual features and attended textual features and feed them into the decoder to predict the coordinates. Table \ref{tab:visual_grounding} demonstrates the performance of mPLUG-\modelname in the visual grounding task. mPLUG-\modelname achieves comparable results with competitive baseline methods.

\begin{figure}[t]

\centering
\subfigure[VQA accuracy.  ]{
\centering
\includegraphics[width=1.6in]{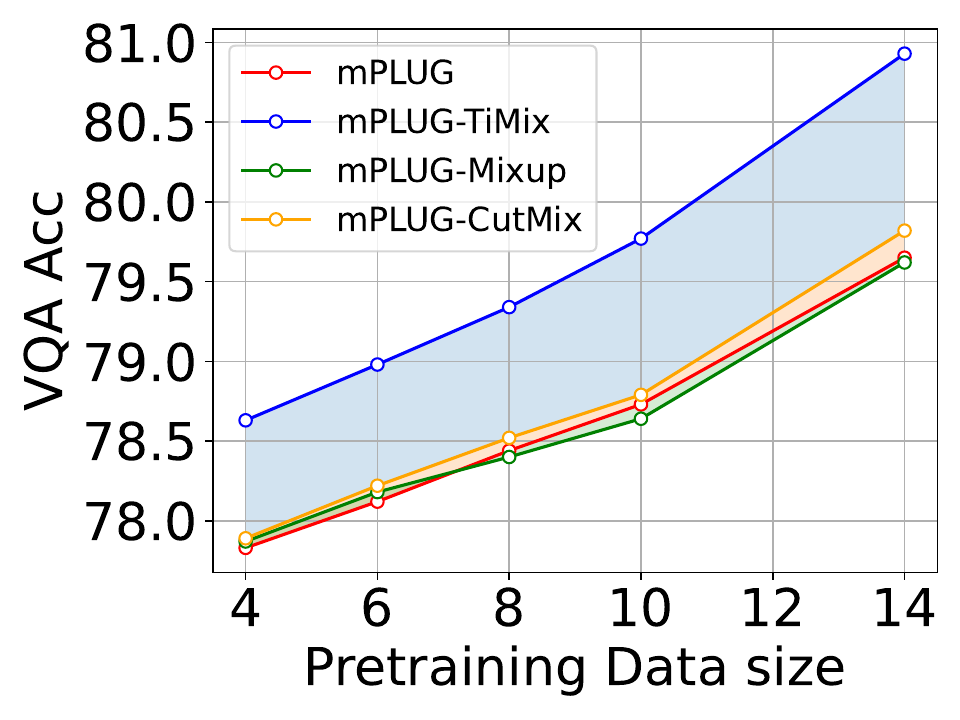}
}%
\subfigure[Pre-training time per epoch ]{
\centering
\includegraphics[width=1.6in]{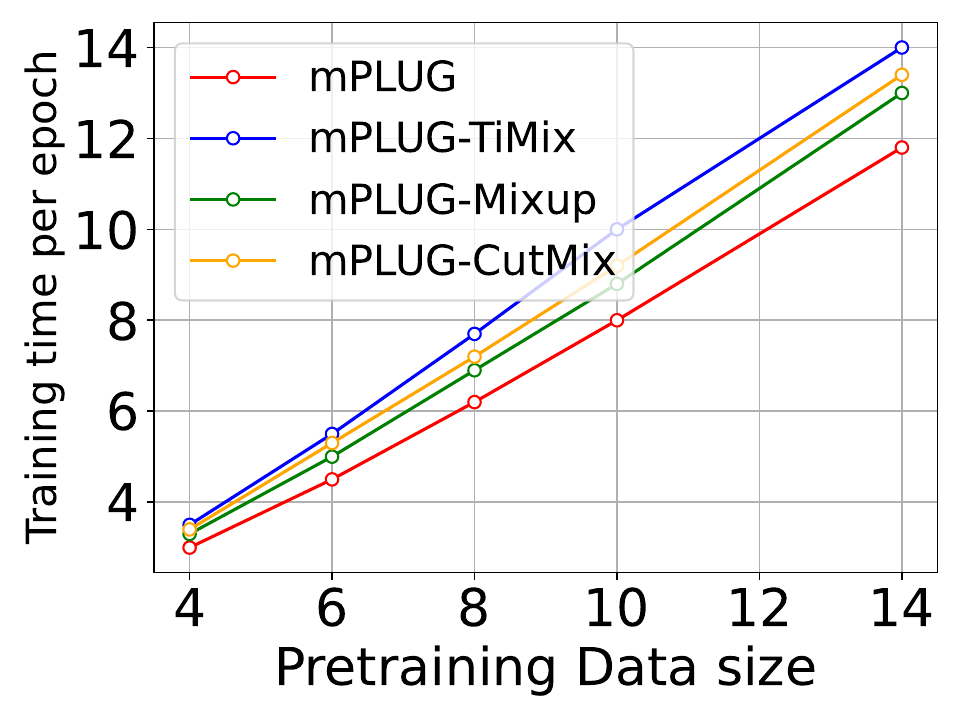}
}
\caption{The visualization of VQA accuracy and Pre-training time per epoch of different models pre-trained on different data sizes}
\label{fig:vqa_acc_datasize}
\end{figure}
\begin{figure}[t]
\centering
\subfigure[ ACC of Text-relevant Patch Prediction]{
\centering
\includegraphics[width=1.6in]{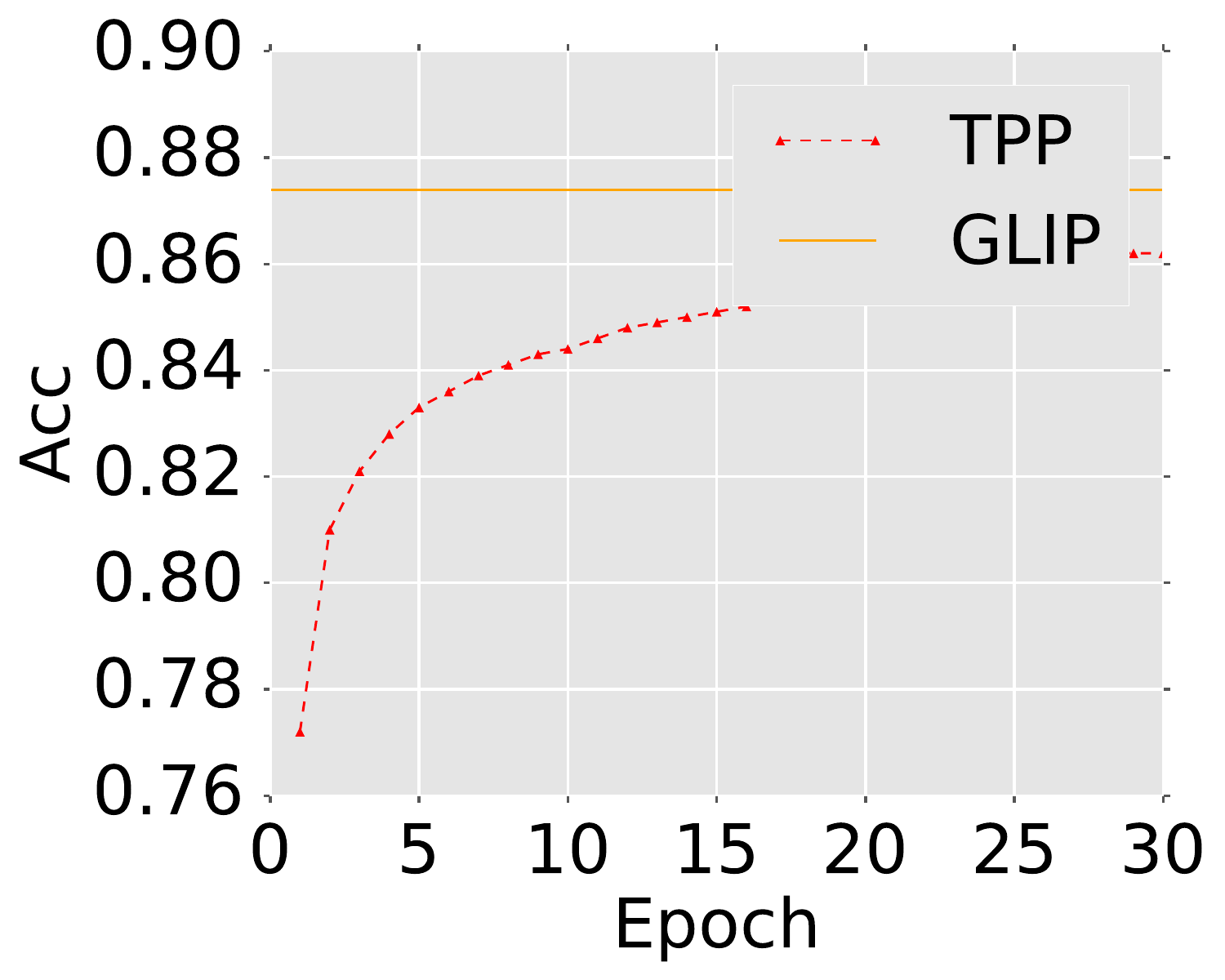}
}%
\subfigure[  Recall of Text-relevant Patch Prediction]{
\centering
\includegraphics[width=1.5in]{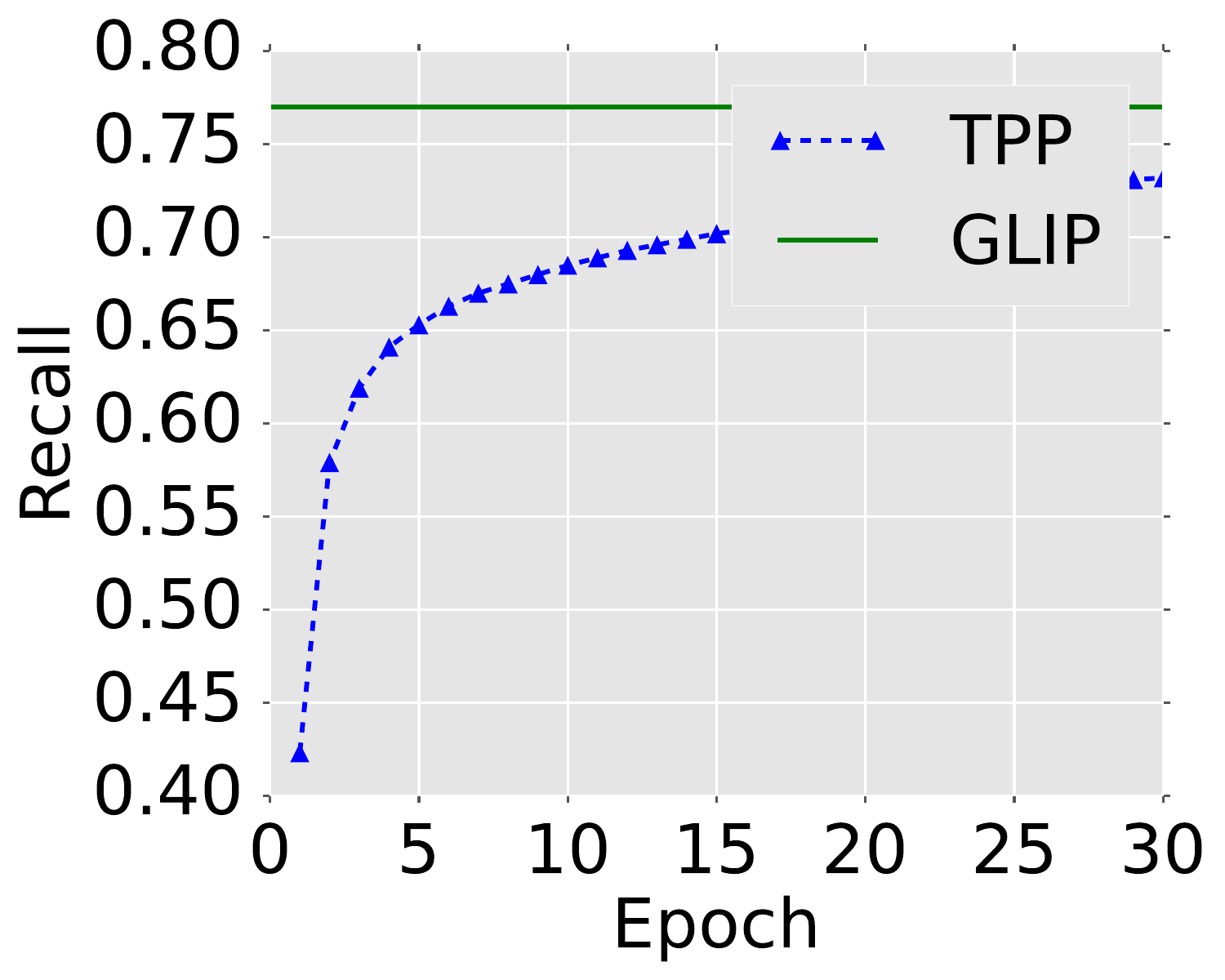}
}%
\caption{The visualization of Accuracy and Recall of TPP on the 10K test dataset randomly sampled from CC~\cite{sharma2018conceptual}.}
\label{fig:acc_recall_TPP}
 
\end{figure}
\subsection{Ablation Study}
 
\begin{table}[t]
\centering
\small
\setlength{\tabcolsep}{2.1mm}{
\begin{tabular}{@{}l|cc|ccc@{}}
\toprule[1.5pt]
model  & PTA & Mix & VQA dev  & NLVR dev  &  PT \\ \midrule
mPLUG-\modelname &   \checkmark   &  \checkmark&  78.63  & 84.12  &  ~3.5h  \\ \hline
\textit{-w/o PTA} &   $\times$   & \checkmark&  77.89  & 82.59  &~3.4h   \\ 
\textit{-w/o Mix} &   \checkmark  & $\times$  &  78.21  &83.13  & ~3.1h     \\ 
\textit{-w/o TiMix} & $\times$   &  $\times$  & 77.83  & 82.66 &~3h \\ 

\bottomrule[1.5pt]
\end{tabular}}
 
\caption{The results of the ablation studies where we report the VQA and NLVR performance of various model variants. PT refers to Pretraining Time }
\label{table6}
\end{table}

 We conducted ablation studies to examine the impact of the Pretraining task Patch Text Alignment (PTA) and mix-based data augmentation. Specifically, we investigated the effects of removing the PTA task while keeping the mix-based data augmentation for contrastive learning (\textbf{w/o PTA}). Without the PTA task, the Text-aware Patch Predictor cannot be optimized effectively, so we replaced it with a simple strategy where we follow the way of CutMix\cite{Yun2019CutMixRS} and randomly sample the region in the image. In Table \ref{table6}, we can observe that without the text guidance (\textbf{w/o PTA}), randomly mixing the image regions leads to a negligible improvement in accuracy on VQA and NLVR compared to the baseline model mPLUG (\textbf{~w/o TiMix}). This demonstrates the effectiveness of our TiMix approach in leveraging text guidance for improved performance. In the case denoted as \textbf{w/o Mix}, we exclude the mix-based data augmentation method and only retain the PTA task. As presented in Table \ref{table6}, we observed that utilizing only the PTA task still leads to a notable improvement in performance. This finding suggests that PTA enables our model to learn fine-grained cross-modal semantic alignment, thereby enhancing performance, although the improvement may not be substantial.

\subsection{Impact of Pre-training Data }

To gain a deeper comprehension of the influence of pre-training data size on the efficacy of \modelname, we conducted pre-training with data sizes of 4M, 6M, 8M, 10M, and 14M. These datasets were further augmented using two different mixing strategies: \modelname, Mixup~\cite{Zhang2017mixupBE} and CutMix~\cite{Yun2019CutMixRS}. Figure \ref{fig:vqa_acc_datasize} (a) showcases the VQA results for various mix strategies, as well as the baseline model mPLUG, which does not employ mix-based augmentation. Notably, \modelname consistently exhibits superior performance across the entire range. This observation suggests that \modelname not only enhances data efficiency in scenarios with limited data but also delivers substantial performance gains as the dataset size expands. From Figure \ref{fig:vqa_acc_datasize} (a), we have observed that CutMix and Mixup provides only limited improvements in model performance. This indicates that the conventional Mixup approach does not significantly enhance the model's performance. Furthermore, it demonstrates the effectiveness of our approach.

\subsection{Data Efficiency of \modelname}

To explore the effects of TiMix on the additional computational costs during pre-training, we conducted experiments to measure the training time per epoch for mPLUG trained without any data augmentation, as well as mPLUG utilizing TiMix, CutMix \cite{Yun2019CutMixRS} and Mixup \cite{Yun2019CutMixRS} for contrastive learning with additional data augmentation. As shown in Figure~\ref{fig:vqa_acc_datasize} (b), we recorded the corresponding training times for various data scales. We found that although \modelname, CutMix and mixup introduce some additional training time and computational overhead, the increase in overhead is not significant. Compared to the baseline model mPLUG, mPLUG-TiMix achieves significant improvements in model performance with relatively less computational cost. For example, to achieve the same performance as mPLUG-TiMix on 4M data size, the baseline mPLUG would require scaling the pre-training data to 10M, resulting in much higher computational overhead. This demonstrates the data efficiency of \modelname. Additionally, as shown in Figure~\ref{fig:gpu training time} (b), we visualize the training loss curves of mPLUG-\modelname trained on 4M data compared to mPLUG trained without TiMix on the same 4M data. We observe that TiMix helps in faster and lower convergence of the training loss. For example, at 30 epochs, the mPLUG-TiMix model has a loss of around 2.1, while the mPLUG model has a loss of around 3.0.

\subsection{Generalization of Text-aware Patch Predictor}

\noindent The text-aware patch predictor (TPP) was trained using the COCO/VG datasets, while our VLP model was trained on more large-scale web-based datasets (e.g., CC\cite{cc12m}/SBU\cite{Ordonez2011Im2TextDI}) having different data distributions with COCO~\cite{Lin2014MicrosoftCC}/VG~\cite{Krishna2016VisualGC}.  Our following experiments aim to evaluate TTP's generalization on potentially out-of-domain data samples e.g., in CC/SBU. In detail, we randomly sample 10K image-text pairs from CC~\cite{sharma2018conceptual} dataset, which is crawled from the web and potentially contains out-of-domain data. As there are no golden bounding box labels existing for the CC dataset. We employ the recent SOTA open-set object detection model, Grounding DINO~\cite{Liu2023GroundingDM}, to detect regions in each image corresponding to the text. Following the approach described in subsection \ref{subsec:PTA}, we convert these regions into patch-level labels. We subsequently evaluate the accuracy and recall of TPP in detecting patches on this test dataset and compared it with the result of GLIP~\cite{glip}, which is also a SOTA open-set grounding model. As depicted in Figure~\ref{fig:acc_recall_TPP}, the accuracy and recall of TPP's patch predictions gradually improve with each training epoch. By the 30th epoch, TPP's accuracy reaches 0.86, and its recall approaches 0.74 which is close to the results of GLIP. These results suggest that TPP is effective and robust in detecting text-related image patches, regardless of whether the image-text pairs originate from manual annotations or are crawled from the internet.

   

\section*{Conclusion}
This paper addresses the challenges of scaling up training data volume in Self-supervised Multi-modal Contrastive Learning (SMCL) for Vision-Language Pre-training (VLP) models. We have introduced Text-aware Image Mixing (\modelname) as a solution to improve data efficiency in VLP by integrating mix-based data augmentation techniques into SMCL. Through a theoretical analysis from a mutual information (MI) perspective, we have theoretically shown that well-mixed data samples serve as a regularizer for the classical InfoNCE loss, empirically resulting in significant performance improvements without incurring excessive computational overhead and thereby significantly improving data efficiency in VLP.

\bibliography{aaai24}

\clearpage
\appendix

\onecolumn
\section{Proof}
\subsection{Proof A}
\label{proof A}
We rewrite the proof provided by ~\citet{oord2018cpc} in the context of image-to-text contrastive learning, where $v_i$ represents an image anchor and $t_i$ and $t_j$ are positive and negative samples, respectively.
 
\begin{align}
& \mathcal{L}_{InfoNCE}^v \nonumber \\
&= -\mathop{E}\limits_{t}\mathop{log} 
                                \left[
                                \frac{ 
                                    \frac{\mathop{P}\left( t_i \middle| v_i \right)}
                                         {\mathop{P}\left( t_i\right)}
                                }
                                { 
                                    \frac{\mathop{P}\left( t_i \middle| v_i \right)}
                                         {\mathop{P}\left( t_i\right)} 
                                     + \sum\limits_{t_j \neq t_i} 
                                        \frac{\mathop{P}\left( t_j \middle| v_i \right)}
                                             {\mathop{P}\left( t_j \right)} 
                                }
                                \right] \\
                                &=\mathop{E}\limits_{t}log 
                                \left[1 + 
                                    \frac{
                                        \mathop{P}\left( t_i\right)
                                    }{
                                        \mathop{P}\left( t_i \middle| v_i \right)
                                    }
                                    \sum\limits_{t_j \neq t_i} 
                                    \frac
                                    {\mathop{P}\left( t_j \middle| v_i \right)}
                                    {\mathop{P}\left( t_j \right)}
                                \right] \\
                                &\approx \mathop{E}\limits_{t}log 
                                \left[
                                  1 + 
                                    \frac{
                                        \mathop{P}\left( t_i\right)
                                    }{
                                        \mathop{P}\left( t_i \middle| v_i \right)
                                    }
                                    \left( N-1 \right)
                                    \mathop{E}\limits_{t_j}
                                        \frac
                                        {\mathop{P}\left( t_j \middle| v_i \right)}
                                        {\mathop{P}\left( t_j \right)}
                                \right] \label{eq12}\\
                                &= \mathop{E}\limits_{t}log 
                                \left[
                                  1 + 
                                    \frac{
                                        \mathop{P}\left( t_i\right)
                                    }{
                                        \mathop{P}\left( t_i \middle| v_i \right)
                                    }
                                    \left( N-1 \right)
                                \right] \label{eq13}\\
                                &\geq \mathop{E}\limits_{t}log 
                                \left[
                                    \frac{
                                        \mathop{P}\left( t_i\right)
                                    }{
                                        \mathop{P}\left( t_i | v_i \right)
                                    } N 
                                \right]\\
                                &= -\mathop{I}(t_i, v_i) + log\left(N\right)    
\end{align}      
Therefore, we have $\mathop{I}(t_i, v_i) \geq log\left(N\right) -\mathcal{L}_{InfoNCE}^v $, where N is the number of batch size.

\subsection{Proof B}
\label{proof B}
We rewrite the \modelname loss based on the \citep{oord2018cpc} in the context of image-to-text contrastive learning, where $\hat{v}^{x,y}$ represents the representation of image anchor mixed from the image $I^x$ and $I^y$ , $t^x$ and $t^y$ are positive and negative samples as follow:
\scriptsize
\begin{align}
&  \mathcal{L}_{\modelname}^v = -\sum\limits_{i = 1:\mathcal{N}} \frac{1}{\mathcal{N}} \left[ \mathbf{s}^x \mathop{log} 
                                \left[
                                \frac{
                                    f\left( \hat{v}^{x,y}_i, t^x_i \right)
                                }
                                {
                                    f\left(  \hat{v}^{x,y}_i, t^x_i\right) + \sum\limits_{t_k \neq t^x_i} 
                                    f\left( \hat{v}^{x,y}_i , t_k\right)
                                 }
                                \right]
                                +
                                \mathbf{s}^y
                                \mathop{log} 
                                \left[
                                \frac{
                                     f\left( \hat{v}^{x,y}_i, t^y_i \right)
                                }
                                {
                                    f\left(  \hat{v}^{x,y}_i, t^y\right) + \sum\limits_{t_k \neq t^y_i} 
                                    f\left(  \hat{v}^{x,y}_i, t_k\right)
                                 }
                                \right]
                                \right]
\nonumber \\
&= -\mathop{E}\limits_{t} \mathbf{s}^x  \mathop{log} 
                                \left[
                                 \frac{ 
                                   \frac{\mathop{P}\left( t^x \middle| \hat{v}^{x,y} \right)}
                                         {\mathop{P}\left( t^x\right)}
                                }
                                { 
                                    \frac{\mathop{P}\left( t^x \middle| \hat{v}^{x,y} \right)}
                                         {\mathop{P}\left( t^x\right)} 
                                      +
                                    \sum\limits_{k \neq i }
                                        \frac{\mathop{P}\left( t_k \middle| \hat{v}^{x,y}\right)}
                                             {\mathop{P}\left( t_k \right)} 
                                }
                                \right]
                                -\mathop{E}\limits_{t}
                                \mathbf{s}^y
                                \left[
                                  \frac{ 
                                   \frac{\mathop{P}\left( t^y \middle| \hat{v}^{x,y} \right)}
                                         {\mathop{P}\left( t^y\right)}
                                }
                                { 
                                    \frac{\mathop{P}\left( t^y \middle|\hat{v}^{x,y} \right)}
                                         {\mathop{P}\left( t^y\right)} +
                                     \sum\limits_{k \neq j }
                                        \frac{\mathop{P}\left( t_k \middle| \hat{v}^{x,y}\right)}
                                             {\mathop{P}\left( t_k \right)} 
                                } \right]  \label{eq1}\\
             & =
                                \mathop{E}\limits_{t} \mathbf{s}^x log 
                                \left[  
                                  1 + 
                                    \frac{
                                        \mathop{P}\left( t^x\right)
                                    }{
                                        \mathop{P}\left( t^x \middle| \hat{v}^{x,y} \right)
                                    }
                                    \sum\limits_{t_k \neq t^x} 
                                    \frac
                                    {\mathop{P}\left( t_k \middle| \hat{v}^{x,y} \right)}
                                    {\mathop{P}\left( t_k \right)}
                                \right]
                                +
                                \mathop{E}\limits_{t} \mathbf{s}^y log 
                                \left[  1 + 
                                    \frac{
                                        \mathop{P}\left( t^y\right)
                                    }{
                                        \mathop{P}\left( t^x \middle| \hat{v}^{x,y} \right)
                                    }
                                    \sum\limits_{t_k \neq t^y} 
                                    \frac
                                    {\mathop{P}\left( t_k \middle| \hat{v}^{x,y} \right)}
                                    {\mathop{P}\left( t_k \right)}
                                \right] \\
                    &\approx \mathop{E}\limits_{t}  \mathbf{s}^x log 
                                \left[
                                  1 + 
                                    \frac{
                                        \mathop{P}\left( t^x\right)
                                    }{
                                        \mathop{P}\left( t^x \middle| \hat{v}^{x,y} \right)
                                    }
                                    \left( N-1 \right)
                                    \mathop{E}\limits_{t_k \neq t^x}
                                        \frac
                                        {\mathop{P}\left( t_k \middle| \hat{v}^{x,y} \right)}
                                        {\mathop{P}\left( t_k \right)}
                                \right] 
                                +
                                \mathop{E}\limits_{t}  \mathbf{s}^y log 
                                \left[
                                  1 + 
                                    \frac{
                                        \mathop{P}\left( t^y\right)
                                    }{
                                        \mathop{P}\left( t^y \middle| \hat{v}^{x,y} \right)
                                    }
                                    \left( N-1 \right)
                                    \mathop{E}\limits_{t_k \neq t^y}
                                        \frac
                                        {\mathop{P}\left( t_k \middle| \hat{v}^{x,y} \right)}
                                        {\mathop{P}\left( t_k \right)}
                                \right] \\
                         &\geq\mathop{E}\limits_{t} \mathbf{s}^x log 
                                \left[
                                  1 + 
                                    \frac{
                                        \mathop{P}\left( t^x\right)
                                    }{
                                        \mathop{P}\left( t^x \middle| \hat{v}^{x,y} \right)
                                    }
                                    \left( N-1 \right)
                                \right] 
                                +
                                 \mathop{E}\limits_{t} \mathbf{s}^y log 
                                \left[
                                  1 + 
                                    \frac{
                                        \mathop{P}\left( t^y\right)
                                    }{
                                        \mathop{P}\left( t^y \middle| \hat{v}^{x,y} \right)
                                    }
                                    \left( N-1 \right)
                                \right] 
                                \label{eq5}\\
                                &\geq \mathop{E}\limits_{t} \mathbf{s}^x log 
                                \left[
                                    \frac{
                                        \mathop{P}\left( t^x\right)
                                    }{
                                        \mathop{P}\left( t^x | \hat{v}^{x,y} \right)
                                    } N 
                                \right] 
                                +
                                \mathop{E}\limits_{t} \mathbf{s}^y log 
                                \left[
                                    \frac{
                                        \mathop{P}\left( t^y\right)
                                    }{
                                        \mathop{P}\left( t^y | \hat{v}^{x,y} \right)
                                    } N 
                                \right]
                                \\
                                &= -\mathbf{s}^x\mathop{I}(t^x, \hat{v}^{x,y}) - \mathbf{s}^y\mathop{I}(t^y, \hat{v}^{x,y}) + (\mathbf{s}^x+ \mathbf{s}^y)log\left(N\right)    \label{eq:eq6}
\end{align}
\normalsize

\noindent which means that:
 \scriptsize
 \begin{align}
 &\mathbf{s}^x\mathop{I}(t^x, \hat{v}^{x,y}) + \mathbf{s}^y\mathop{I}(t^y, \hat{v}^{x,y})  \geq   (\mathbf{s}^x+ \mathbf{s}^y)log\left(N\right) -\mathcal{L}_{\modelname}^v \label{sup:eq27}
 \end{align}
 \normalsize
Noted that  $\hat{v}^{x,y} $ is mixed from image $v^x$ and $v^y$.  Thus, we have $\hat{v}^{x,y} = (v_{\textbf{r}}^x,v_{\textbf{r}}^y) $. 
 \scriptsize
\begin{align}
    & \mathop{I}(t^x, \hat{v}^{x,y}) = \mathop{I}(t^x, (v_{\textbf{r}}^x, v_{\textbf{r}}^y)) \label{sup:eq:1}\\
    & \mathop{I}(t^y, \hat{v}^{x,y}) = \mathop{I}(t^y, (v_{\textbf{r}}^x, v_{\textbf{r}}^y)) \label{sup:eq:2}
\end{align}
\normalsize
According to the Chain Rule for MI, we have:
\scriptsize
\begin{align}
    & \mathop{I}(t^x, (v_{\textbf{r}}^x, v_{\textbf{r}}^y)) = \mathop{I}(t^x, v_{\textbf{r}}^x) +\mathop{I}(t^x, 
    v_{\textbf{r}}^y|v_{\textbf{r}}^x) \label{eq:2}
\end{align}
\normalsize
If $v_{\textbf{r}}^y$ and $v_{\textbf{r}}^x$ are not independent and contain overlapping information about $t^x$. In this case, we have :
\scriptsize
\begin{align}
    & \mathop{I}(t^x, v_{\textbf{r}}^x) +\mathop{I}(t^x, v_{\textbf{r}}^y)> \mathop{I}(t^x, v_{\textbf{r}}^x) +\mathop{I}(t^x, v_{\textbf{r}}^y|v_{\textbf{r}}^x)  \label{eq:3}
\end{align}
\normalsize
Otherwise if $v_{\textbf{r}}^y$  and $v_{\textbf{r}}^x$ are independent, we have:
\scriptsize
\begin{align}
    & \mathop{I}(t^x, v_{\textbf{r}}^x) +\mathop{I}(t^x, v_{\textbf{r}}^y) = \mathop{I}(t^x, v_{\textbf{r}}^x) +\mathop{I}(t^x, v_{\textbf{r}}^y|v_{\textbf{r}}^x)  \label{eq:4}
\end{align}
\normalsize
Overall, combine equations~\ref{eq:2},~\ref{sup:eq:1}, ~\ref{eq:4} and inequality~\ref{eq:3}, we have:
\scriptsize
\begin{align}
     & \mathop{I}(t^x, v_{\textbf{r}}^x) +\mathop{I}(t^x, v_{\textbf{r}}^y) \geq  \mathop{I}(t^x, \hat{v}^{x,y}) \label{eq:31}
\end{align}
\normalsize
Similarly, we also have:
\scriptsize
\begin{align}
     & \mathop{I}(t^y, v_{\textbf{r}}^x) +\mathop{I}(t^y, v_{\textbf{r}}^y) \geq  \mathop{I}(t^y, \hat{v}^{x,y}) \label{eq:32}
\end{align}
\normalsize

\noindent Therefore,  combining inequality \ref{sup:eq27}, \ref{eq:32}, \ref{eq:31}, we have:
\scriptsize
\begin{align}
     &\mathbf{s}^x (\mathop{I}(t^x, v_{\textbf{r}}^x) + \mathop{I}(t^x, v_{\textbf{r}}^y)) +  \mathbf{s}^y(\mathop{I}(t^y, v_{\textbf{r}}^x) + \mathop{I}(t^y, v_{\textbf{r}}^y)) \nonumber  \\
     & \geq   log\left(N \right)   - \mathcal{L}_{\modelname}^v  \label{eq:9}
\end{align}
\normalsize
which equal to: 
\scriptsize
\begin{align}
     &\mathbf{s}^x \mathop{I}(t^x, v_{\textbf{r}}^x) +  \mathbf{s}^y\mathop{I}(t^y, v_{\textbf{r}}^y) \nonumber  \\
     & \geq   log\left(N \right) - (\mathcal{L}_{\modelname}^v + \mathbf{s}^x\mathop{I}(t^x, v_{\textbf{r}}^y)+\mathbf{s}^y \mathop{I}(t^y, v_{\textbf{r}}^x) ) 
     \label{eq:eqIM3}
\end{align}
\normalsize


\twocolumn
\section{Additional Experiments}
\subsection{Impact of \modelname for Contrastive Learning and Modality Collaboration }

\begin{figure}[htbp]
\centering
\subfigure[ UMAP visualization w/o \modelname]{
\centering
\includegraphics[width=1.4in]{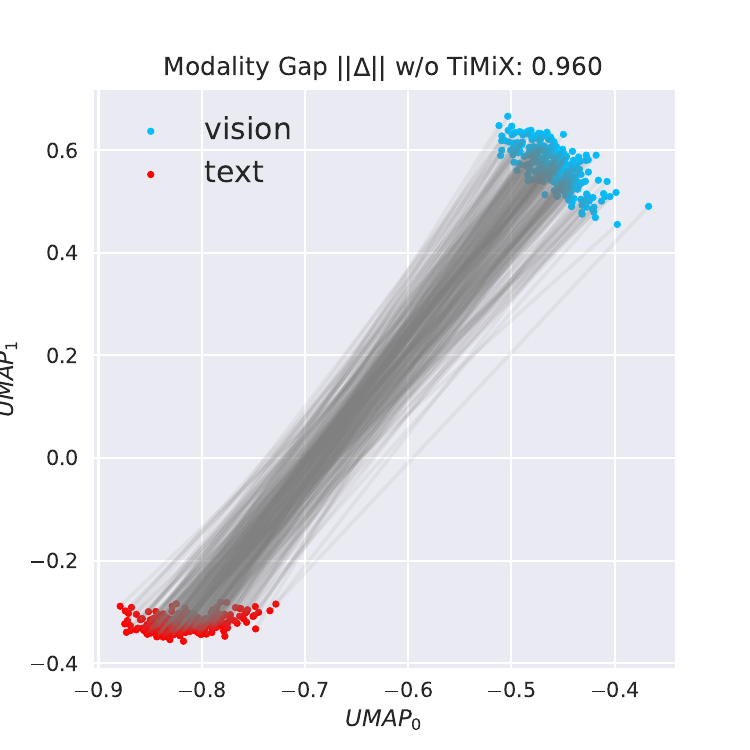}
}%
\subfigure[ UMAP visualization with \modelname]{
\centering
\includegraphics[width=1.4in]{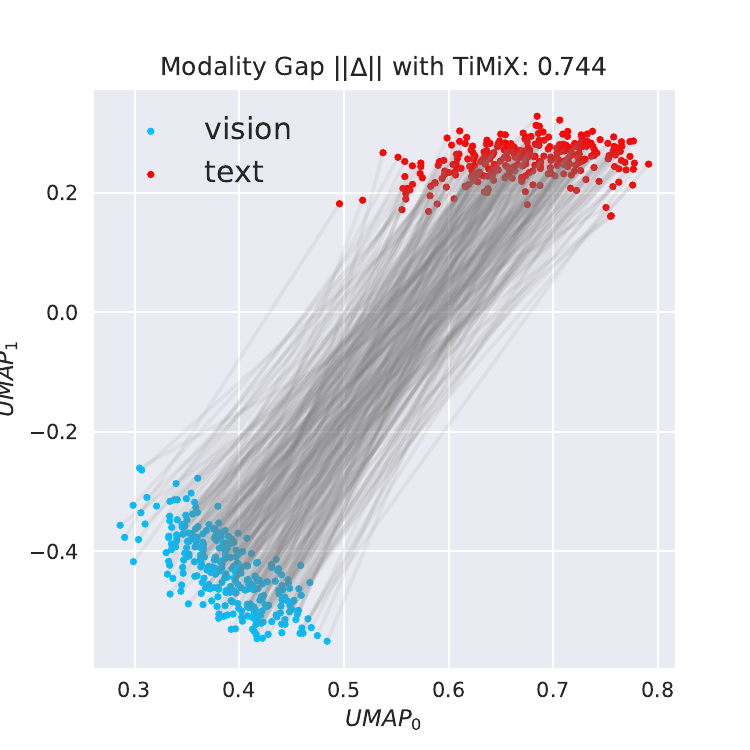}
}
\caption{The UMAP visualization of generated vision and language embeddings from mPLUG with and w/o \modelname. The black lines refer to vision-language pairs.}
\label{fig:umap visualization}
\end{figure}

Here we investigate the influence of \modelname in terms of the representation space learned by contrastive Learning and the modality collaboration.  We randomly sample some image-text pairs, and sketch the UMAP visualization of the generated embeddings from pre-trained mPLUG-TiMix in Figure~\ref{fig:umap visualization}. 
We can observe that with the assistance of \modelname, the distance between vision and text samples is closer and  the distribution of samples is more uniform.  In contrast, models without \modelname tend to have more concentrated embedding distributions. Besides, we quantitatively compute the modality gap \cite{Liang2022MindTG}, where $\Vert \Delta \Vert $ is the difference between the center of vision embeddings and text embeddings.  It can be observed that the \modelname would encourage the collaboration between vision and language modalities thus yielding a lower modality gap compared to the model without \modelname ( 0.96 $\Vert \Delta \Vert $ without \modelname vs  0.744 $\Vert \Delta \Vert $ with \modelname).

\section{Experiment Settings}
\label{sup:setting}

\subsection{Implementation Details}
We pre-train ALBEF-\modelname and mPLUG-\modelname for 30 epochs with a total batch size of 1024 on 8 NVIDIA A100 GPUs.  We use the AdamW optimizer~\citep{Loshchilov2019DecoupledWD} with a weight decay of 0.02 to train our models. The learning rate is warmed up to 1e-5 (ViT-B/16) and 1e-4 (BERT$_{base}$) in the first 1000 iterations. During pre-training, we use input images with a resolution of $256 \times 256$ and increase the image resolution during fine-tuning. For ALBEF\modelname and mPLUG-\modelname, we utilize a 6-layer Transformer for both the text encoder and the cross-modal fusion network. Following \cite{Li2021AlignBF}, we initialize the text encoder with the first 6 layers of BERT$_{base}$~\citep{Devlin2019BERTPO} and the cross-modal network with the last 6 layers of BERT$_{base}$. For image-text contrastive learning, we set the queue size to 65,536 and the momentum coefficient to 0.995.

\subsection{pre-training data}
\label{sup: data}

Following the previous work~\cite{Li2021AlignBF}, for 4M data size, we use the same pre-training dataset which includes two in-domain datasets (MS COCO ~\cite{Lin2014MicrosoftCC} and Visual Genome ~\cite{Krishna2016VisualGC}), and
three web out-domain datasets (Conceptual Captions~\cite{sharma2018conceptual}, SBU Captions ~\cite{Ordonez2011Im2TextDI}, for 14M data size, we also use same the pretraining dataset which includes MS COCO ~\cite{Lin2014MicrosoftCC} and Visual Genome ~\cite{Krishna2016VisualGC}, and 
two web two-domain datasets, Conceptual 12M ~\cite{cc12m}, SBU Captions ~\cite{Ordonez2011Im2TextDI}. 

\begin{table}[htbp]
\setlength\tabcolsep{3pt}
\centering
\small
\begin{tabular}{l|ccccc}
\toprule[1.5pt]
  &  COCO & VG & SBU & CC3M & CC12M \\
\midrule
image & 113K & 100K & 860K & 3M & 12M \\
text & 567K & 769K & 860K & 3M  & 12M \\
\bottomrule[1.5pt]
\end{tabular} 
\caption{Statistics of the pre-training datasets.}
\label{table:pre-traindata}
\end{table}

\begin{table}[htbp]
\setlength\tabcolsep{10.3pt}
\centering

\begin{tabular}{l|cccc}
\toprule[1.5pt]
     &  image & Captions & Objects & Regions \\
\midrule
COCO & 0.11M & 0.55M & 0.45M & -   \\
VG   & 0.10M & - & 2.0M & 3.7M \\
\bottomrule[1.5pt]
\end{tabular} 
\caption{Statistics of objects/regions annotations used in the pre-training.}
\label{table:objectdata}
\end{table}

Table \ref{table:pre-traindata} shows the statistics of the 4M images with texts used in the pre-training stage. Besides, as shown in Table~\ref{table:objectdata} we use also use the objects/regions annotations from COCO\cite{Lin2014MicrosoftCC} and VG \cite{Krishna2016VisualGC} datasets, and we give statistics of object and region annotations of each dataset. Note that we use the object/region annotations provided by ~\citet{Zeng2021xvlm} thus we follow their setting which filtered out some samples because of: 1) invalid annotations (e.g. negative values for bounding boxes or boxes being outside of the images); 2) boxes being too small (< 1\%); 3) highly overlapped textual descriptions of regions (>75\%), etc. After pre-processing,  we keep COCO objects 446,873 (from 859,999), VG objects 2,043,927 (from 3,802,349), VG regions 3,699,598 (from 5,402,953).

\subsection{Pre-training Schedule}
As shown in the Algorithm~\ref{alg:pre-train}. After calculating the \PretrainTaskName loss $\mathcal{L}_{PTA}$, we then randomly sample a mini-batch of normal image-text pairs from the 4M or 12M dataset and calculate the Image-Text Contrastive (ITC) loss $\mathcal{L}_{ITC}$, Image-Text Matching (ITM) loss $\mathcal{L}_{ITM}$, Masked Language Modeling (MLM) loss $\mathcal{L}_{MLM}$ and Prefix Language Modeling (PrefixLM) loss $\mathcal{L}_{Prefix}$ based on other four pre-training objectives. Then, we use \modelname to augment the mini-batch data and get the mixed data. We calculate the contrastive loss $\mathcal{L}_{\modelname}^v$ and $\mathcal{L}_{\modelname}^t$ based on the mixed mini-batch data. We assign equal loss weights to each pre-training loss, and thus the full pre-training loss is:
\begin{equation}
     \mathcal{L} = \mathcal{L}_{ITC}+\mathcal{L}_{ITM} +\mathcal{L}_{MLM} + \mathcal{L}_{Prefix} + \mathcal{L}_{PTA}+ \mathcal{L}_{\modelname}^v + \mathcal{L}_{\modelname}^t
\end{equation}
Besides, at the beginning of pre-training, as the PTA loss has not yet converged, thus the performance of the patch predictor is not ideal, we will not apply \modelname during the first 2 epochs. Thus in the first 2 epochs, the full pre-training loss is:
\begin{equation}
     \mathcal{L} = \mathcal{L}_{ITC}+\mathcal{L}_{ITM} +\mathcal{L}_{MLM} + \mathcal{L}_{Prefix} + \mathcal{L}_{PTA}
\end{equation}  
As the PTA loss gradually converges, we will use the patch predictor to detect the text-relevant patches in the SPD module.

\begin{algorithm*}[htbp]

  \caption{Pre-training of TiMix}
  \label{alg:pre-train}

  \KwIn{Large scale pre-training dataset $\mathcal{D}$, Object/Region Dataset $\mathcal{O}$, the number of pre-training epochs $T$, the pre-training learning rate $\alpha$, the batch size $B_D$ of dataset $\mathcal{D}$, the batch size $B_O$ of dataset $\mathcal{O}$.  }

  Initialize the parameters $\theta$ of our model $M$ \;
  
  \For{$t=1$ to $T$}{
    Randomly sample a mini-batch of $B_O$ Images $\{\hat{v}_1, \hat{v}_2, \dots, \hat{v}_{B_O} \}$ from $\mathcal{D}$ \;
    
    \For{$i=1$ to $B_O$}{
      Select a object or region $r_i$ from image $\hat{v}_i$ \;
      
      Translate the object class label $\hat{y}_i$ to text description $\hat{t}_i$\;
      
      Translate the bounding box annotation of $r_i$ to patch annotations $ Y^i = \{y^i_1,y^i_2, \dots, y^i_n\}$ \;
     
    }
    
    Run forward of $M$ on the mini-batch of image-text pairs $\{\{\hat{v}_1, \hat{t}_1\}, \{\hat{v}_2, \hat{t}_2\}, \dots, \{\hat{v}_{B_O}, \hat{t}_{B_O}\} \}$ and $\{Y^1, Y^2, \dots, Y^{B_O}\}$ to obtain the loss $\mathcal{L}_{PTM}$ \;

    Randomly sample a mini-batch of $B$ Image-Text Pairs $\{\{v_1,t_1\}, \{v_2,t_3\}, \ldots, \{v_{B_D},t_{B_D}\}\}$ from $\mathcal{D}$ \;

    Run forward of $M$ on the mini-batch of image-text pairs $\{\{v_1,t_1\}, \{v_2,t_3\}, \ldots, \{v_{B_D},t_{B_D}\}\}$  to obtain the losses $\mathcal{L}_{ITC}$, $\mathcal{L}_{ITM}$, $\mathcal{L}_{MLM}$, $\mathcal{L}_{Prefix}$ \;
    
    Use \modelname to augment the mini-batch of $B$ Image-Text Pairs $ \{\{v_1,t_1\}, \{v_2,t_3\}, \ldots, \{v_{B_D},t_{B_D}\}\}$ to the mixed mini-batch of $B$ Image-Text Pairs $\{\{\hat{v}^{x,y}_1,t^x_1,t^y_1\}, \{\hat{v}^{x,y}_2,t^x_2,t^y_2\}, \ldots, \{\hat{v}^{x,y}_{B_D},t^x_{B_D},t^y_{B_D}\}\}$;
    
   Run forward of $M$ on the mini-batch of image-text pairs  $\{\{\hat{v}^{x,y}_1,t^x_1,t^y_1\}, \{\hat{v}^{x,y}_2,t^x_2,t^y_2\}, \ldots, \{\hat{v}^{x,y}_{B_D},t^x_{B_D},t^y_{B_D}\}\}$ to obtain the losses  $\mathcal{L}_{\modelname}^v$ and $\mathcal{L}_{\modelname}^t$
    
    Calculate the overall loss:
    
    $\mathcal{L} = \mathcal{L}_{ITC}+\mathcal{L}_{ITM} +\mathcal{L}_{MLM} + \mathcal{L}_{Prefix} + \mathcal{L}_{PTM}$ + $\mathcal{L}_{\modelname}^v$ +$\mathcal{L}_{\modelname}^t$ \;
    
    Backward the overall loss $\mathcal{L}$ and update the parameters of $M$ using gradient descent with learning rate $\alpha$ and the average loss $\mathcal{L}$ over the mini-batch:

    $\theta \leftarrow \theta - \alpha \frac{1}{B} \sum_{i=1}^{B} \nabla_{\theta} \mathcal{L}(\theta; s_i)$ \;

  }

  \Return $M$ with pre-trained parameters $\theta$ \;

\end{algorithm*}

\subsection{Pre-training Objectives}
\label{pre-traing objectives}
We pre-train our model with five standard objectives: \PretrainTaskName (PTA), Image-Text Contrastive learning (ITC), Image-Text Matching (ITM), and  Masked Language Modeling (MLM),Prefix Language Modeling (PrefixLM). These pre-training tasks are optimized jointly. As we have talked the \PretrainTaskName before, in this subsection, we only introduce the last four pre-training task.

\noindent \textbf{Image-text Contrastive (ITC)} For \modelname, We follow the \cite{Li2021AlignBF} and apply ITC to align the image representation and text representation from the unimodal encoders. For the image, the image feature corresponding to the image [CLS] token is chosen as the image representation. For the text, the text token feature corresponding to the text [CLS] token is the text representation.

\noindent \textbf{Image-Text Matching (ITM)}  The goal of image-text matching is to predict whether the input image and text are matched.  We follow the design of \cite{Li2021AlignBF} and select hard negative image-text pairs based on the contrastive text-image similarity. We take the text [CLS] embedding of the multimodal encoder's output as the joint representation, followed by a Multi-Layer Perceptron (MLP) layer for prediction.

\noindent \textbf{Masked Language Modeling (MLM)} The task setup is basically the same as in BERT~\cite{Devlin2019BERTPO}, where we randomly mask 15$\%$ of tokens in text and the model is asked to predict these masked words with the cross-modal representations.

\noindent \textbf{Prefix Language Modeling (PrefixLM).} This task aims to generate the caption given an image and predict the text segment subsequent to the cross-modal context as ~\cite{bi2020palm}. It optimizes a cross entropy loss by maximizing the likelihood of text in an autoregressive manner.

\section{Downstream Task Details}
\label{sup:downsteam}

We evaluate \modelname on the four downstream vision-language tasks. The hyperparameters that we use for finetuning on the downstream tasks are listed in Table \ref{table:finetune-hyper}. Following ~\citep{Li2021AlignBF}, all tasks adopt RandAugment, AdamW optimizer with a weight decay of 0.05 and a cosine learning rate schedule. Next, we introduce the dataset settings in detail.


\paragraph{VQA.} The VQA task ~\cite{Agrawal2015VQAVQ} requires the model to answer natural language questions given an image. Most methods~\cite{Tan2019LXMERTLC,Wang2021VLMoUV,Li2020OscarOA,Wang2021SimVLMSV} deal with visual question-answering tasks as multi-label classification on pre-defined answer sets. This strategy achieves strong performance, but it is not suitable for real-world open scenarios. We conduct an experiment on the VQA2.0 dataset ~\citep{goyal2017making}, which contains 83k/41k/81k images for training/validation/test. Following ~\citep{Li2021AlignBF}, we use both training and validation splits for training, and incorporate additional training data from Visual Genome~\citep{Krishna2016VisualGC}. Besides, we concatenate the question with the object labels and OCR tokens extracted from the image.
\begin{table}
\setlength\tabcolsep{2pt}
\centering

\begin{tabular}{l|ccc}
\toprule
Task  &  LR (ViT-B/BERT ) & batch size & epochs  \\
\midrule
VQA & 2e-5/5e-6 & 1024 &  8 \\
Captioning$\dagger$ & 1e-5\&8e-7 & 256& 5 \\
Retrieval & 1e-5/2e-6 & 256& 5 \\
Visual Grounding & 2e-5/2e-6 & 512& 120 \\
\bottomrule
\end{tabular} 
\caption{Finetuning hyperparameters for downstream tasks. $\dagger$ denotes two-stage fine-tuning.}
\label{table:finetune-hyper}
\end{table}
\paragraph{Image Captioning.} The image captioning task requires a model to generate an appropriate and fluent caption for a given image. We evaluate image captioning on two datasets COCO Caption~\cite{Lin2014MicrosoftCC} and NoCaps~\cite{nocaps}. \modelname finetuned with training data of COCO Caption is tested on both of the datasets. We train \modelname on the MS COCO Caption and test on the same Karpathy split~\cite{Li2020OscarOA,Wang2021SimVLMSV} and NoCaps validation set. Following~\cite{Li2020OscarOA}, we first fine-tune \modelname with the cross-entropy loss for 5 epochs with a learning rate of 1e-5 and a batch size of 256. Based on the fine-tuned model, we then fine-tune it with CIDEr optimization~\cite{scst} for extra 5 epochs with a smaller learning rate of 8e-7.  We use the best checkpoint on COCO Caption and predict on the Nocaps validation set directly. During inference, we use beam search with a beam size of 10 and set the maximum generation length as 20.

\paragraph{Image-Text Retrieval.} We conduct experiments for both image-to-text retrieval (TR) and text-to-image retrieval (IR) on COCO ~\cite{Lin2014MicrosoftCC} and Flickr30K ~\cite{Plummer2015Flickr30kEC} datasets. We adopt the widely-used Karpathy split ~\citep{karpathy2015deep} for both COCO and Flickr30K. COCO contains 113k/5k/5k images for train/validation/test, and Flickr30K contains 29k/1k/1k images for train/validation/test. Following ~\cite{Li2021AlignBF, li2022blip}, we jointly optimize the ITC loss and the ITM loss during fine-tuning. During inference, we first select top-k candidates by computing the dot-product similarity between the image and text encoder features and then rerank the selected candidates based on their ITM scores (In the fine-grained reranking stage, for the same image, we re-extracting multiple image encoder features based on TPP with the guidance of multiple text candidates.). We set $k = 256$ for COCO and $k = 128$ for Flickr30K. 

\paragraph{NLVR2.} The NLVR2~\citep{Suhr2019ACF} task requires the model to predict whether a sentence. We conduct experiments following the original train/val/test split in ~\citep{Suhr2019ACF}. Following \citep{li2022blip}, we use two cross-attention layers to process the two input images, and their outputs are merged and fed to the FFN. An MLP classifier is then applied to the output embedding of the language [CLS] token.

\paragraph{Visual Grounding.} The task of visual grounding involves localizing the referred object in an image given a plain text query. Instead of directly regressing bounding boxes, our approach concatenates visual features with textual features, which are then fed into the multi-modal decoder to predict the object's coordinates. We evaluate our method on the referring expression grounding dataset: RefCOCO+\cite{yu2016modeling}. The RefCOCO+ dataset contains 19K images and 141K queries.

\section{Comparison Models}
\label{sup:comparison models}

\begin{itemize}

\item \textbf{E2E-VLP}~\citep{Xu2021E2EVLPEV}: proposes the first end-to-end VLP method for both V+L understanding and generation, with a unified Transformer encoder-decoder architecture.

\item \textbf{VinVL}~\citep{2021VinVL}: pre-trains a large-scale object-attribute detection model with much larger amounts of supervised data on four public object detection datasets for extracting better region-based visual features. 

\item \textbf{OSCAR}~\citep{Li2020OscarOA}: proposes to use object tags detected in images as anchor points to ease the learning of cross-modal alignments, where the input to the Transformer is a combination of image, text, and object tags.

\item \textbf{METER}~\citep{dou2021empirical}: systematically investigates how to design and pre-train a fully transformer-based VL model in an end-to-end manner.

\item \textbf{VLMo}~\citep{Wang2021VLMoUV}: presents a unified vision-language pre-trained model that jointly learns a dual encoder and a fusion encoder with a modular Transformer network.

\item \textbf{SimVLM}~\citep{Wang2021SimVLMSV}: different from previous VLP methods that only use limited (4M-10M) image-text pairs for pre-training, it proposes a simple VLP model with a single prefix language modeling objective, which pre-trains on an extremely large aligned cross-modal data of about 1.8B noisy image-text pairs. This is also the latest state-of-the-art method of image captioning. 

\item \textbf{ALBEF}~\citep{Li2021AlignBF}: introduces a contrastive loss to align the image and text representations before fusing them through cross-modal attention, which enables more grounded vision and language representation learning.

\item \textbf{UNITER}~\citep{Chen2020UNITERUI}: proposes an improved single-stream VLP method, by designing two new pre-training strategies: 1) it uses conditional masking on pre-training tasks instead of random masking strategy, 2) it designs a new word-region alignment pre-training task via the use of optimal transport to explicitly encourage fine-grained alignment between words and image regions. 



\item  \textbf{ViLT}~\citep{Kim2021ViLTVT}:  adopts linear projection and word embedding as the visual and textual encoders, and uses the visual transformer as the cross-modal encoder to align and fuse the features of both modalities in an end-to-end manner.

\item \textbf{BLIP}~\citep{li2022blip}: proposes a new VLP framework that transfers flexibly to both vision-language understanding and generation tasks. It effectively utilizes noisy web data by bootstrapping the captions.
\item \textbf{METER}~\cite{dou2021empirical}: systematically investigates how to design and pre-train a fully transformer-based VL model in an end-to-end manner.

\item \textbf{LXMERT}~\citep{Tan2019LXMERTLC}: is the pioneering work to pre-train a dual-stream multi-modal Transformer, which consists of an object relationship encoder, a language encoder, and a cross-modality encoder.

\item \textbf{ViLBERT}~\citep{Lu2019ViLBERTPT}: proposes the first work that extends the BERT architecture to a multi-modal dual-stream VLP model, which processes both visual and textual inputs in separate streams that interact through co-attentional transformer layers.

\item \textbf{mPLUG} ~\citep{li2022mplug}: is a  vision-language foundation model for both cross-modal understanding and generation and introduces an effective and efficient vision-language architecture with novel cross-modal skip-connections.

\item \textbf{XVLM}~\cite{Zeng2021xvlm}: proposes to learn multi-grained alignments which locate visual concepts in the image given the associated texts, and in the meantime align the texts with the visual concepts.

\item \textbf{SCL}~\cite{scl}: proposes the semantic completion learning task to  facilitate global-to-local cross-modal alignment which obtains state-of-the-art performance on various vision-language benchmarks.

\item \textbf{MAP}~\cite{map}: studies the modeling of multimodal semantic uncertainty and project the representations of all modalities as probabilistic distributions via a probability distribution encoder by utilizing sequence-level interactions.

\end{itemize}

\section{More Visualization}

\begin{figure*}
    \centering
    \includegraphics[width=5.5in]{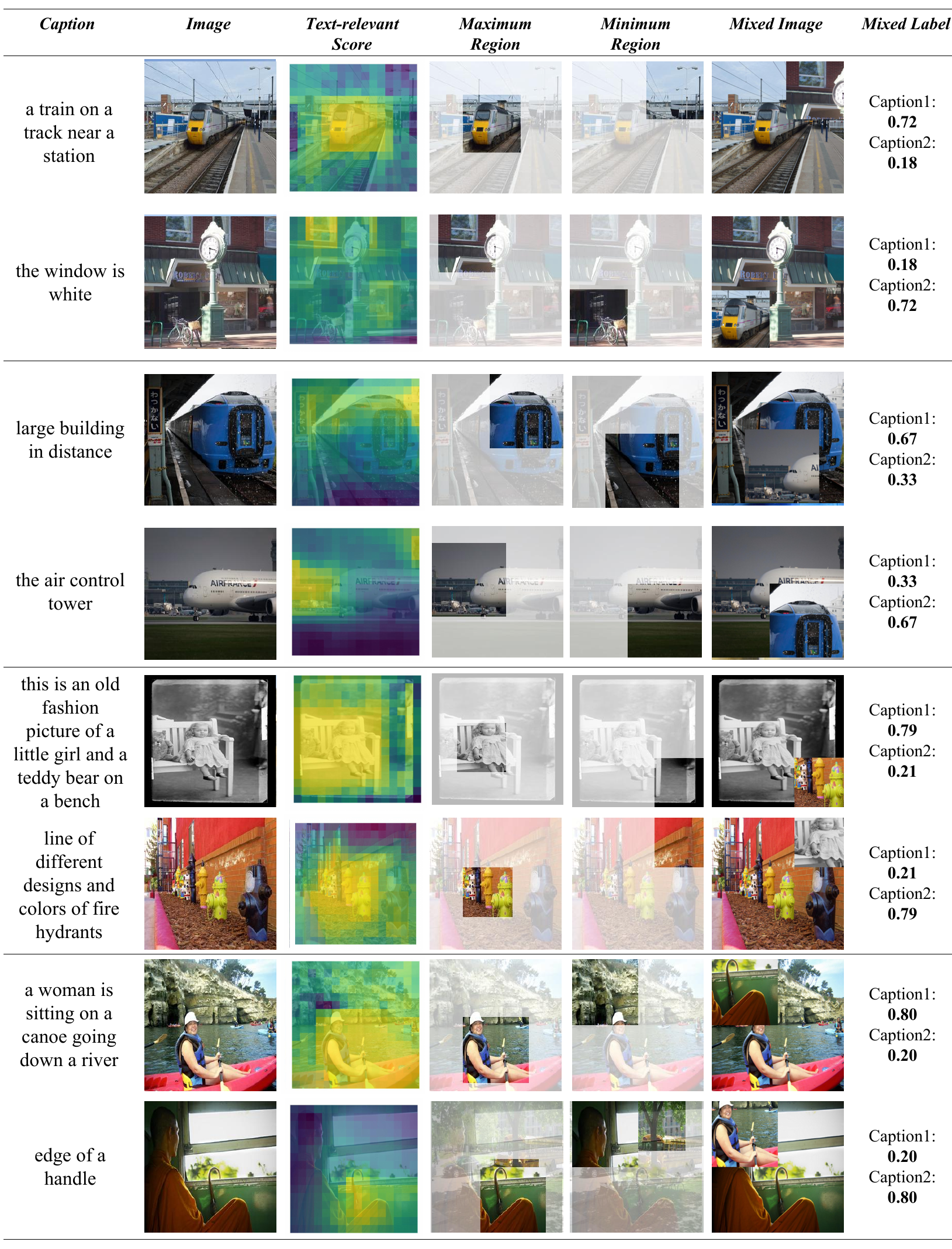}
    \caption{Visualization of examples enhanced with TiMix data augmentation. }
    \label{fig:case3}
\end{figure*}

\begin{figure*}
    \centering
    \includegraphics[width=5.5in]{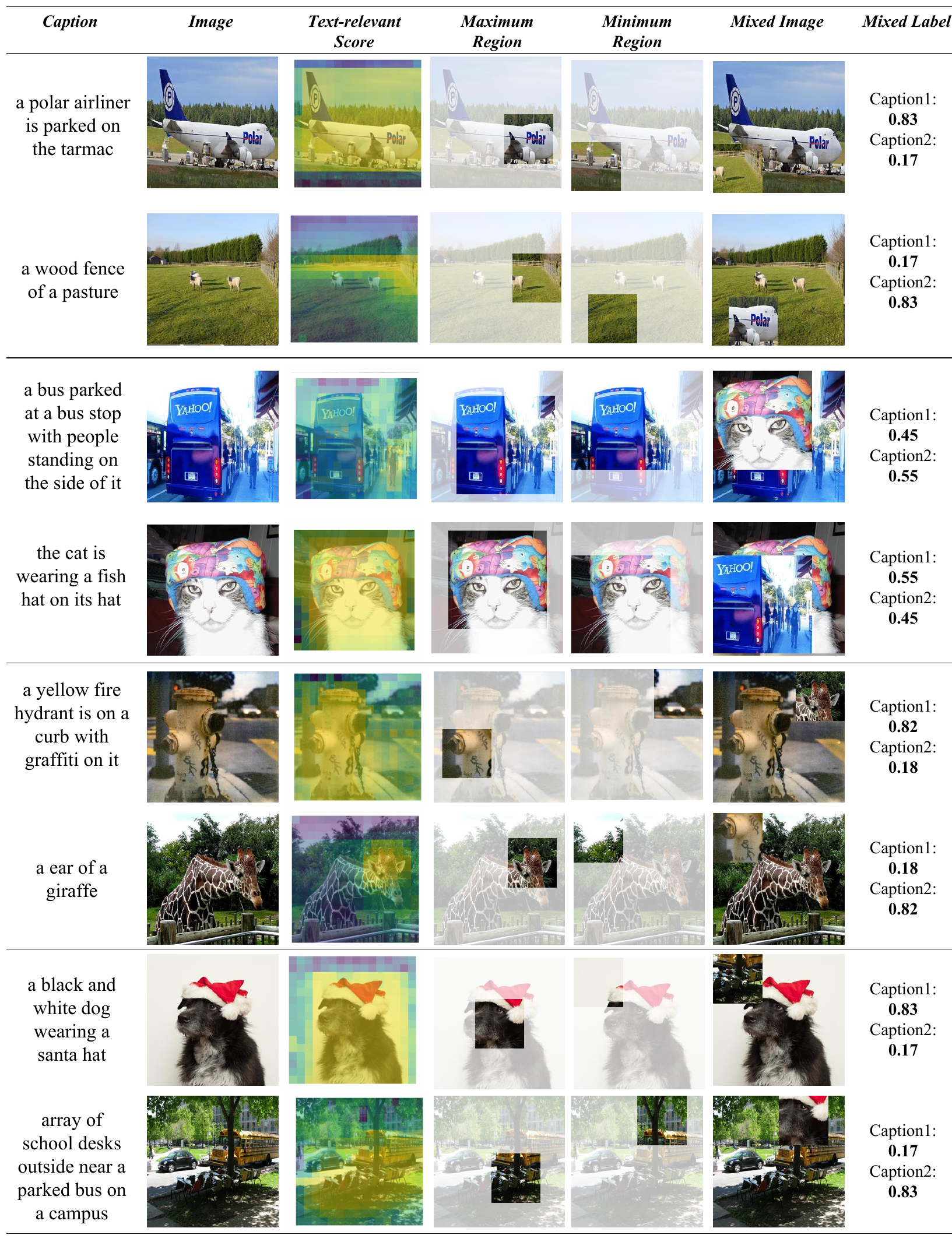}
    \caption{Visualization of examples enhanced with TiMix data augmentation. }
    \label{fig:case1}
\end{figure*}

\begin{figure*}
    \centering
    \includegraphics[width=5.5in]{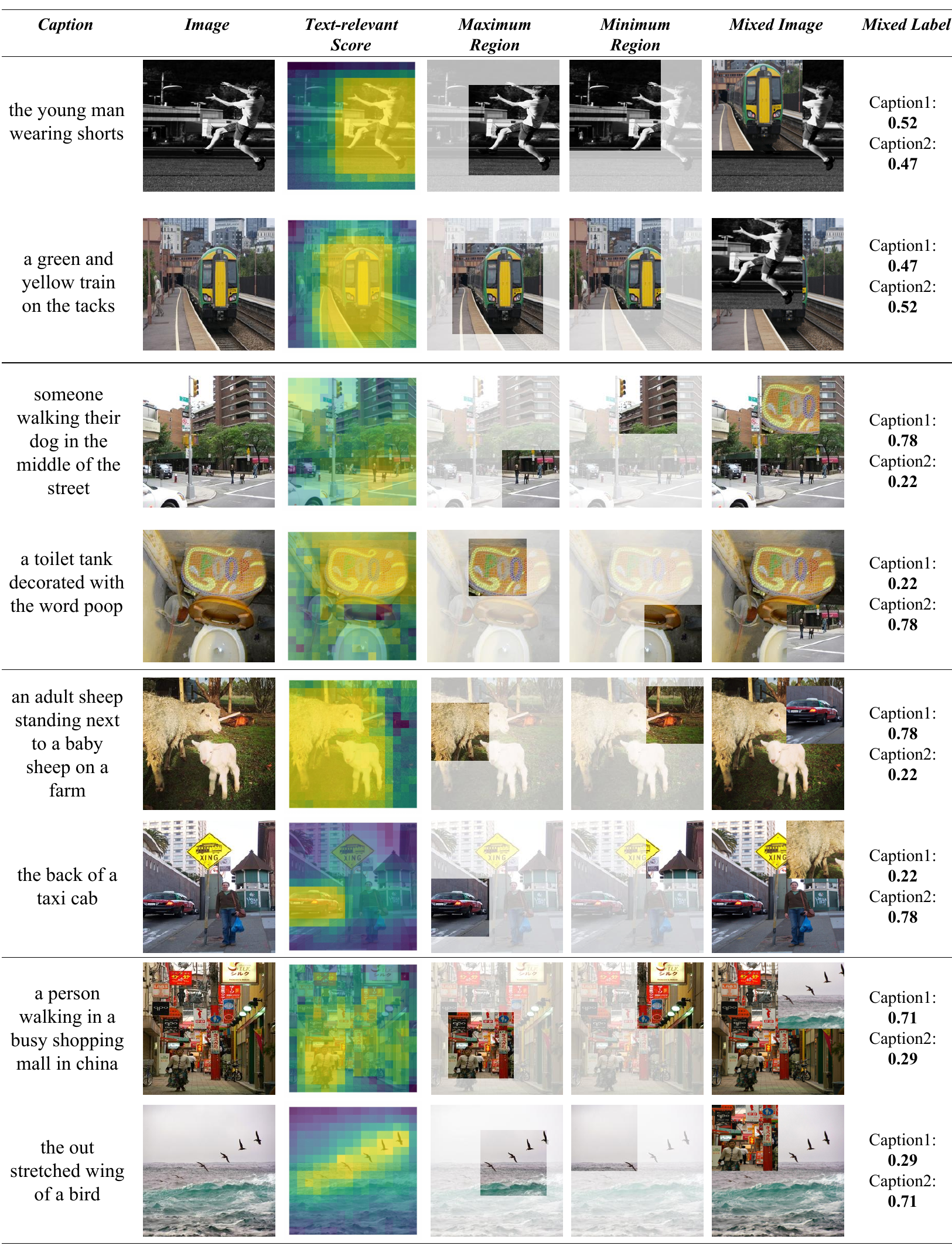}
    \caption{Visualization of examples enhanced with TiMix data augmentation. }
    \label{fig:case2}
\end{figure*}

In this section, we present additional visual analysis cases to provide a further illustration. Please refer to Figure~\ref{fig:case3}, Figure~\ref{fig:case2} and Figure~\ref{fig:case1} for the visualizations.

\clearpage

\end{document}